\documentclass[10pt,twocolumn,letterpaper]{article}

\usepackage{cvpr}              % To produce the CAMERA-READY version

\usepackage{multicol}
\usepackage[toc,page]{appendix}
\usepackage{soul}
\usepackage{amsmath}

\definecolor{cvprblue}{rgb}{0.21,0.49,0.74}
\usepackage[pagebackref,breaklinks,colorlinks,allcolors=cvprblue]{hyperref}

\def\paperID{47} % *** Enter the Paper ID here
\def\confName{CVPR}
\def\confYear{2025}

\title{No Train Yet Gain:\\Towards Generic Multi-Object Tracking in Sports and Beyond}

\author{
Tomasz Stanczyk$^{1,2}$ \qquad Seongro Yoon$^{1,2}$ \qquad Francois Bremond$^{1,2}$\\
$^1$ Inria, France  \hspace{0.5cm} $^2$ Université Côte d’Azur, France\\
{\tt\small tomasz.stanczyk@inria.fr, seong-ro.yoon@inria.fr, francois.bremond@inria.fr}
}

\begin{document}
\maketitle
% \input{sec/0_abstract}    
% \input{sec/1_intro}
% \input{sec/2_formatting}
% \input{sec/3_finalcopy}

%%%%%%%%% ABSTRACT
\begin{abstract}
   % \textbf{\textcolor{red}{[TO DO!]}} 
   % Multi-object tracking (MOT) involves identifying and consistently tracking objects across video sequences. Traditional tracking-by-detection methods, while effective, often require extensive tuning and lack generalizability. On the other hand, segmentation mask-based methods are more generic but struggle with tracking management, making them unsuitable for MOT. We propose a novel approach, McByte, which incorporates a temporally propagated segmentation mask as a strong association cue within a tracking-by-detection framework. By combining bounding box and propagated mask information, McByte enhances robustness and generalizability without per-sequence tuning. Evaluated on four benchmark datasets - SportsMOT, DanceTrack, SoccerNet-tracking 2022 and MOT17 - McByte demonstrates performance gain in all cases examined. At the same time, it outperforms existing mask-based methods. Implementation code will be provided upon acceptance.
   Multi-object tracking (MOT) is essential for sports analytics, enabling performance evaluation and tactical insights. However, tracking in sports is challenging due to fast movements, occlusions, and camera shifts. Traditional tracking-by-detection methods require extensive tuning, while segmentation-based approaches struggle with track processing.
   % management. 
   We propose McByte, a tracking-by-detection framework that integrates temporally propagated segmentation mask as an association cue to improve robustness without per-video tuning. Unlike many existing methods, McByte does not require training, relying solely on pre-trained models and object detectors commonly used in the community. Evaluated on SportsMOT, DanceTrack, SoccerNet-tracking 2022 and MOT17, McByte demonstrates strong performance across sports and general pedestrian tracking. Our results highlight the benefits of mask propagation for a more adaptable and generalizable MOT approach. Code will be made available at \url{https://github.com/tstanczyk95/McByte}.
\end{abstract}

%%% --- THAT'S THE GOOD ONE (Second visual exemple)! --- %%%
% \begin{figure*}
% \centering
% \begin{tabular}{cccc}
% \includegraphics[height=4.0cm]{images/bt2a.png}&
% \includegraphics[height=4.0cm]{images/bt2b.png}&
% \includegraphics[height=4.0cm]{images/ours2a.png}&
% \includegraphics[height=4.0cm]{images/ours2b.png}
% \\
% Frame 459 (baseline)&Frame 475 (baseline)&Frame 459 (McByte)&Frame 475 (McByte)
% \end{tabular}
% \vspace*{0.3cm}
% \caption{Visual output comparison between the baseline and McByte. With the temporally propagated mask guidance, McByte can handle the association of an ambiguous set of bounding boxes and reduce the identity switches - see the subjects with IDs 59 and 63 on the output of McByte. Input image data from~\cite{mot17_ref}. Best seen in color.}
% \label{fig:visual_differences_2}
% \end{figure*}
%%% --- </> --- %%%

%%%%%%%%% BODY TEXT
\section{Introduction}
\label{sec:intro}

% --- OLD (v1) ---
% \textbf{\textcolor{red}{"Tr. management"... Tr. Processing?}} Multi-object tracking (MOT) is a computer vision task that involves tracking objects (e.g., people) across video frames while maintaining consistent object IDs. MOT detects objects in each frame and associates them across consecutive frames. Applications include sport analysis, surveillance and autonomous driving, making reliable MOT trackers essential.
%
% --- OLD (v2) ---
% \textbf{\textcolor{red}{[ "Tr. management"... Tr. Processing? ] }} 
% Multi-object tracking (MOT) is a computer vision task that involves tracking objects across video frames while maintaining consistent object IDs. It gets initiated to detect objects in each frame and associate them across consecutive frames. Applications include sport analysis, surveillance and autonomous driving, making reliable MOT trackers essential.
%
% --- NEW ---
Multi-object tracking (MOT) is to involve tracking objects across video frames while maintaining consistent object IDs. It gets initiated to detect objects in each frame and associate them across consecutive frames. MOT can be applied to tracking players and performers in various sport settings to facilitate both team and individual performance analysis and statistics. Nevertheless, these sport settings are still challenging due to dynamic movement of the targets, blurry objects and occlusions, posing strong demands to tracking solutions.

% --- OLD (v1) ---
% Tracking-by-detection methods~\cite{bt_ref,ocsort_ref, deepocsort_ref,cbiou_ref,strongsort_ref,hybridsort_ref} use bounding boxes to detect objects in each frame and associate them with those from previous frames, based on cues like position, appearance, and motion. The resulting matches form "tracklets" over consecutive frames. However, these methods often require extensive hyper-parameter tuning for each dataset or even per single sequence, reducing their generalizability and limiting their application across different datasets and various sport types and settings.
%
% --- NEW ---
Tracking-by-detection~\cite{bt_ref,ocsort_ref, deepocsort_ref,cbiou_ref,strongsort_ref,hybridsort_ref}, one of the intuitive approaches, firstly detects the target objects with bounding boxes in each frame and associates them with those from previous frames, based on cues such as position, appearance, and motion. Then, the resulting matches form "tracklets" over consecutive frames. However, these methods often require extensive hyper-parameter tuning for each dataset or even per single sequence, reducing their generalizability and limiting their application across different datasets and various sport scenarios.

\begin{figure}
\centering
\begin{tabular}{ccc}
\includegraphics[height=2.4cm]{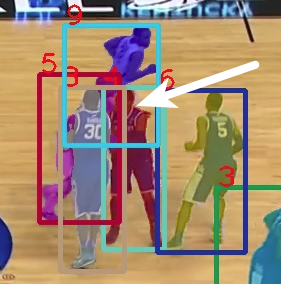}&
\includegraphics[height=2.4cm]{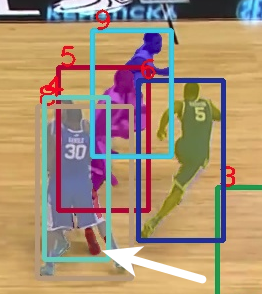}& % &
\includegraphics[height=2.4cm]{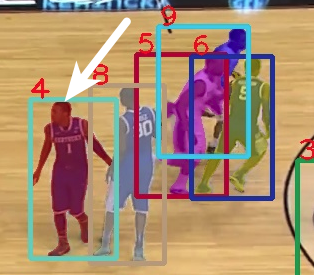}
% (a)&(b)&(c)
\end{tabular}
% \vspace*{0.3cm}
\caption{Mask propagation module can be helpful in cases of severe occlusion. The person with the red mask is tracked only by its limited visible parts (pointed by white arrows), particularly in the middle image. Input image data from~\cite{sportsmot_ref}.}
\vspace*{-0.4cm}
\label{fig:power_of_cutie}
\end{figure}

Segmentation mask-based methods~\cite{deva_ref,sam2_ref}, on the other hand, generate masks to cover objects and track them across video frames. Trained on large datasets, these methods aim to capture the semantics of image patches, making them more generic. However, they are not designed for MOT, lacking robust management for tracking multiple entities and struggling to detect new objects entering the scene, especially team sport players with dynamically moving camera. Additionally, these methods rely entirely on mask predictions for object positioning, which can be problematic when the predictions are noisy or inaccurate.

% --- OLD (v1) ---
% In this paper, we explore applying a temporally propagated segmentation mask as an association cue to assess its effectiveness in MOT, in particular in tracking players and performers in different sports. We propose a novel tracking-by-detection method that combines mask propagation and bounding boxes to improve the association between tracklets and detections. 
% %
% % The mask propagation is managed according to the tracklet lifespan, with usage constraints to enhance tracking performance. 
% %
% The mask propagation is managed according to the tracklet lifespan, while the mask is used in a controlled manner to enhance tracking performance. 
% %
% Since the applied temporal mask propagation model has been trained on a large dataset, it makes the entire tracking process more generic. Unlike existing tracking-by-detection methods, our approach does not require tuning hyperparameters for each dataset or video sequence, i.e. it is not parametric.
%
% --- NEW ---
In this paper, we explore applying temporal mask propagation method as an association cue to assess its effectiveness in MOT of challenging scenarios in sport media. We propose a novel tracking-by-detection method that combines mask propagation module and detections to improve the association. Since the applied temporal mask propagation model has been trained on a large dataset, it makes the entire tracking process more generic. Unlike existing tracking-by-detection methods, our approach does not require sensitive tuning of hyper-parameters for each dataset or video sequence, i.e. it is not parametric.

% \textbf{\textcolor{red}{[To be mentioned in the contributions]}} 
A combination of a linear motion prediction model such as Kalman filter~\cite{kf_ref} and detection matching based on intersection over union (IoU), which is a baseline for most of the tracking-by-detection methods~\cite{ocsort_ref, deepocsort_ref,cbiou_ref,strongsort_ref,hybridsort_ref}, is not suitable on its own for settings with highly dynamic motion and blurry objects, in tracking the sport players or performers. We believe that the temporally propagated mask might be able to handle this. However, providing segmentation mask ground truths for each video frame, also for blurry objects, is a huge burden. In our approach, we do not perform any costly training of the mask segmentation or temporal propagator. Instead, we apply well-studied models pre-trained on a huge corpus of data without specifically tuning them on MOT datasets, but with carefully incorporating them in our method. 

% --- OLD (v1), moved down to "We evaluate our..." ---
% \textbf{\textcolor{blue}{[ MOVE IT DOWN, "We evaluate our..." ]}} Further, for a fair comparison with other methods, we use the same, pre-trained object detectors used in the community. Solely with provided detection, mask segmentation and mask temporal propagation models, and Kalman filter, we obtain good results and satisfactory performance on five different datasets, in particular in sport settings.
% % \textbf{\textcolor{magenta}{[+ something about transfromers and such - that they require a lot of training? And our approach does not!]}} 

% --- OLD (v1) ---
% Tracking multiple objects at once often involves handling challenging occlusions, where only a small part of the subject might be visible, e.g. a leg of a person. Temporally propagated mask can be especially helpful in such cases, when the visible shape can considerably differ from the subject. A visual example is presented in \cref{fig:power_of_cutie}.
% --- NEW ---
Tracking multiple objects at once often involves handling challenging occlusions. To be specific, when many players try to take over the ball, where only a small part of the subject might be visible. Temporally propagated mask can be especially helpful in such cases, when the visible shape can considerably differ from the subject. A visual example is presented in \cref{fig:power_of_cutie}, where person with ID 4 is severely occluded, but the mask (in red) still identifies their shoe and helps to maintain tracking the person.

We note explicitly that incorporating temporally propagated mask, which also involves temporal coherency, is different than using a static mask coming directly from an image segmentation model~\cite{sam_ref} independently per each frame. Using the mask temporal propagation as an association cue within the problem of MOT has not been done before. 
We evaluate our incorporation of the temporally propagated mask as an association cue against a baseline tracker. We show clear benefits for MOT, including (but not limited to) sport settings, by handling challenging situations such as ambiguous occlusions and blurry tracked objects. Our tracker is tested on four MOT datasets, while for a fair comparison, 
% with other methods,
we use the same, pre-trained object detectors used in the community. Our method outperforms tracking-by-detection algorithms on SportsMOT~\cite{sportsmot_ref}, DanceTrack~\cite{dancetrack_ref}, SoccerNet-tracking 2022~\cite{soccernet-tracking2022_ref} and MOT17~\cite{mot17_ref}. These results highlight the advantages of using mask propagation, eliminating the need for per-sequence hyper-parameter tuning. Solely with Kalman filter, provided detection, mask segmentation and 
% mask temporal 
propagation models, we obtain meaningful results and performance on four different datasets.

Our contribution in this work is summarized as follows:
    (i) We design a tracking-by-detection algorithm, based on Kalman filter and a pre-trained mask propagation model, which does not require any expensive training (\textit{no train, yet gain}).
    (ii) We propose a practical idea adapting a temporally propagated object segmentation mask as an effective association cue incorporated into an MOT tracking algorithm. The tracker overcomes the limitations of mask-based approaches by performing proper tracklet management and including other important association cues as well as the limitations of the baseline tracking-by-detection approaches, by making the tracking process more robust and generic.
    (iii) We propose practical policies enforcing regulated usage of the propagated mask in a controlled manner, which help handle common situations in sport settings, such as occlusions and blurry objects.
\begin{figure*}
\centering
\includegraphics
% [width=18cm]
[width=15cm]
% [width=14cm]
{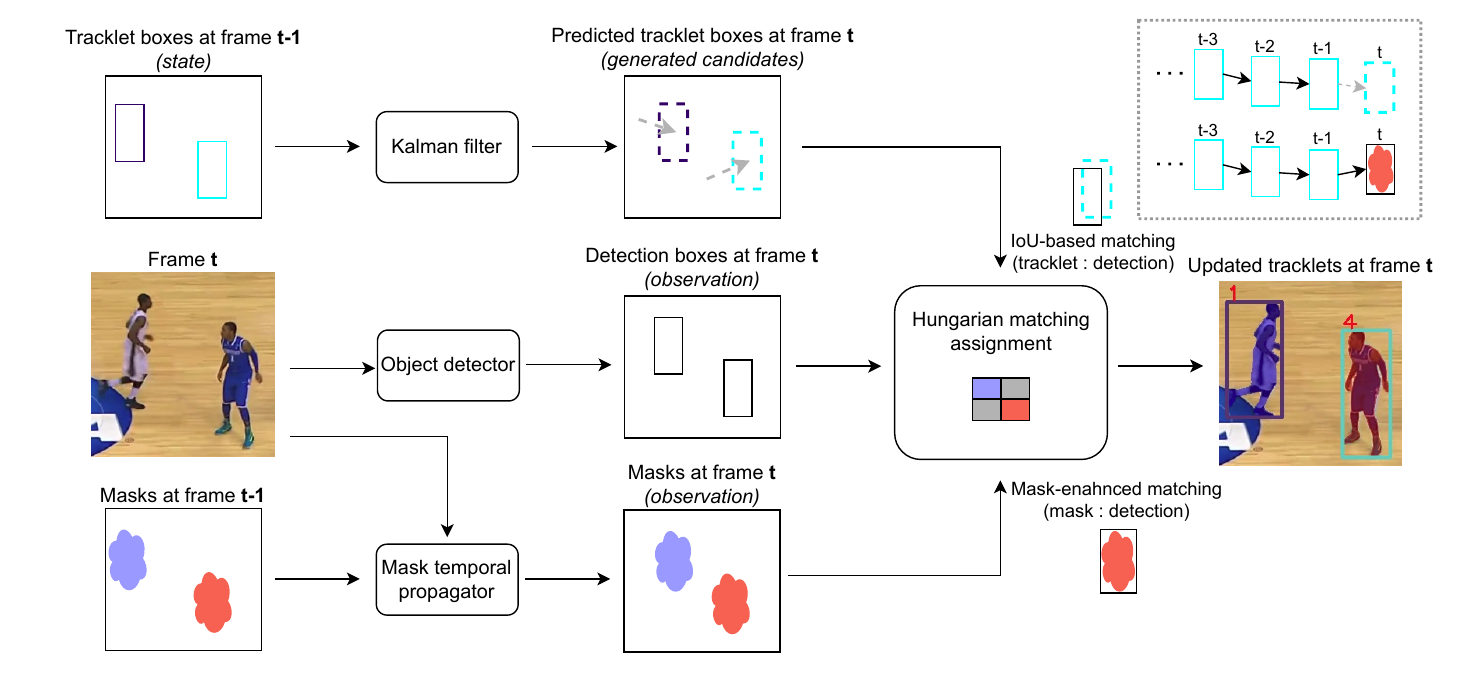}
% {images/CVPRW25_main_diagram__working_copy_1.drawio.png}
% \vspace*{0.3cm}
% \caption{\textbf{\textcolor{red}{[UPDATE THE CAPTION!]}} McByte tracking pipeline with the mask cue guidance. Temporally propagated mask signal is incorporated as an association cue in the tracklet-detection association steps.}
\caption{The overview of our proposed tracking pipeline. \textbf{Top:} Existing tracklet bounding box positions are considered as the tracker state from the previous frame, t-1. Using Kalman filter, candidate tracklet positions are generated and considered as a predicted next state. \textbf{Middle:} Current frame t is passed to object detector. Then, initial association cost matrix is computed based on IoU between the bounding boxes of tracklet candidate positions and detections. \textbf{Bottom:} Mask temporal propagator takes the current frame t as an input and produces the masks, which are then considered as additional observation. The cost matrix is enriched with the information based on the matching between masks and detections. Finally, the cost matrix is passed to the Hungarian matching algorithm minimizing the association cost and the tracklets at frame t are updated with the matched detections.}
% --- Chat GPT... ---
% \caption{\textbf{\textcolor{red}{Is it Good?}} Overview of our tracking pipeline. (Top): Tracklet bounding boxes from frame t-1 serve as the tracker state, with Kalman filter predicting candidate positions. (Middle): The object detector processes frame t, and an initial cost matrix is computed using IoU between tracklet candidates and detections. (Bottom): The mask propagator generates masks for frame t, enriching the cost matrix with mask-detection matching. The final cost matrix is optimized using the Hungarian algorithm, updating tracklets for frame t.}
\vspace*{-0.3cm}
\label{fig:diagram_tracking_pipeline}
\end{figure*}

\section{Related Work}
\label{sec:related_work}

% \noindent\textbf{MOT methods.} Transformer-based methods~\cite{motip_ref,motrv2_ref,memotr_ref, motr_ref} use attention mechanisms to learn tracking trajectories and object associations from training data, following an end-to-end approach. Further, as opposed to our method, transformer-based methods require a lot of training data. Other types of MOT methods which have also been proposed in the community, include global optimization (offline) methods~\cite{sushi_ref} or joint detection and tracking methods~\cite{fairmot_ref,relationtrack_ref}.
% --- Chat GPT ---
% \noindent\textbf{MOT methods.}
% \noindent\textbf{Existing MOT frameworks and limitations.}
\noindent\textbf{Multi-object tracking approaches.}
Transformer-based approaches~\cite{motip_ref,motrv2_ref,memotr_ref,motr_ref} use attention mechanisms for end-to-end learning of tracking trajectories and object associations but require extensive training data. Other MOT methods include global optimization (offline tracking)~\cite{sushi_ref} and joint detection-tracking approaches~\cite{fairmot_ref,relationtrack_ref}.

% Tracking-by-detection approaches detect objects in each video frame and associate them into tracklets, linking the same objects across frames.
% ByteTrack~\cite{bt_ref} is a powerful tracking-by-detection algorithm that uses the YOLOX~\cite{yolox_ref} detector and Kalman Filter~\cite{kf_ref} linear motion model, associating tracklets based on the intersection over union (IoU) between bounding boxes of tracklets and new detections. It provides strong tracklet management and serves as a solid baseline for MOT. Several works have built on ByteTrack. OC-SORT~\cite{ocsort_ref} enhances state estimation by computing virtual trajectories during occlusion. StrongSORT~\cite{strongsort_ref} adds re-identification (re-ID) features as cues, camera motion compensation, and NSA Kalman filter~\cite{nsa_kf_ref}. C-BIoU~\cite{cbiou_ref} extends the association process by buffering (enlarging) bounding boxes, while HybridSORT~\cite{hybridsort_ref} adds cues like confidence modeling and height-modulated IoU alongside existing strong cues.
% --- Chat GPT ---
Tracking-by-detection methods detect objects in each frame and associate them into tracklets. ByteTrack~\cite{bt_ref}, a strong baseline, uses the YOLOX~\cite{yolox_ref} detector and Kalman Filter~\cite{kf_ref} to associate tracklets via intersection over union (IoU). Several methods extend ByteTrack: OC-SORT~\cite{ocsort_ref} improves state estimation with virtual trajectories during occlusion, StrongSORT~\cite{strongsort_ref} incorporates re-identification (re-ID) features, camera motion compensation, and NSA Kalman filter~\cite{nsa_kf_ref}, C-BIoU~\cite{cbiou_ref} enlarges bounding boxes for better association, and HybridSORT~\cite{hybridsort_ref} adds confidence modeling and height-modulated IoU.

% Although these algorithms perform well on popular MOT datasets such as MOT17~\cite{mot17_ref}, they are highly sensitive to parameters. ByteTrack, for instance, explicitly tunes parameters like high-confidence detection 
% % and tracklet initialization 
% thresholds per test sequence in their practical implementation. Extensions of ByteTrack~\cite{strongsort_ref,ocsort_ref,deepocsort_ref, hybridsort_ref} also rely on per-sequence parameter tuning. This tuning process is costly and impractical for larger datasets like SportsMOT~\cite{sportsmot_ref}, DanceTrack~\cite{dancetrack_ref}, and SoccerNet-tracking~\cite{soccernet-tracking2022_ref}, as well as for generalizing across datasets and scene settings, e.g. different sports. In contrast, our method, incorporating the temporally propagated segmentation mask as an association cue, avoids per-sequence tuning, making it more robust and generic. 
% --- Chat GPT ---
Although effective on MOT datasets like MOT17~\cite{mot17_ref}, these algorithms are highly sensitive to parameters. ByteTrack, for example, performs per-sequence tuning of detection thresholds in its practical implementation, and its extensions~\cite{strongsort_ref,ocsort_ref,deepocsort_ref,hybridsort_ref} also rely on extensive parameter adjustments. This process is costly and impractical for large datasets like SportsMOT~\cite{sportsmot_ref}, DanceTrack~\cite{dancetrack_ref}, and SoccerNet-tracking~\cite{soccernet-tracking2022_ref}, as well as for generalizing across different sports. In contrast, our method leverages temporally propagated segmentation masks as an association cue, eliminating the need for per-sequence tuning and improving robustness across diverse settings.
\noindent\textbf{Mask Temporal Propagation.} XMem~\cite{xmem_ref}, based on the Atkinson-Shiffrin Memory Model~\cite{atk_shf_mem_model_ref}, enables long-term segmentation mask tracking in video object segmentation (VOS). Its successor, Cutie~\cite{cutie_ref}, enhances segmentation by incorporating object encoding from mask memory for better background differentiation. While image segmentation models like SAM~\cite{sam_ref} generate initial masks at specific frames, mask temporal propagation models infer and update masks across subsequent frames.

% % While useful for re-identifying subjects across the adjacent frames, both XMem and Cutie are not suitable for MOT as they do not involve bounding boxes and can provide inaccurate mask predictions as listed in their performance on video object segmentation~\cite{xmem_ref, cutie_ref}. 
% While generally powerful in their target domains, both XMem and Cutie are not directly suitable for MOT, as they do not involve bounding boxes and might still provide inaccurate mask predictions as listed in their performance on video object segmentation~\cite{xmem_ref, cutie_ref}. 
% Therefore, we propose a novel MOT algorithm that combines temporally propagated masks with bounding boxes, improving tracking performance.
% --- Chat GPT ---
Although effective in their domains, XMem and Cutie are not directly suited for MOT as they do not involve bounding boxes and can produce inaccurate mask predictions~\cite{xmem_ref, cutie_ref}. To address this, we propose a novel MOT algorithm that integrates temporally propagated masks with bounding boxes, enhancing tracking performance.
\noindent\textbf{Mask-Based Tracking Systems.} Segmentation mask models have been applied to tracking. DEVA~\cite{deva_ref}, an extension of XMem~\cite{xmem_ref}, introduces decoupled video segmentation and bi-directional propagation, involving masks and bounding boxes. Grounded SAM 2 combines Grounding Dino~\cite{gr_dino_ref} and SAM 2~\cite{sam2_ref} for bounding box tracking and object ID maintenance. MASA~\cite{masa_ref}, a SAM-based~\cite{sam_ref} mask feature adapter~\cite{adapters_ref,adapters_cv_ref}, supports video segmentation and object tracking by matching detections across frames. However, these mask-based approaches lack robust tracklet management and struggle with occlusions, new object entries, and missed detections.

\section{Proposed method}
\label{sec:method}

\subsection{Preliminaries}
\label{sec:method_preliminaries}

In tracking-by-detection methods~\cite{bt_ref,ocsort_ref, deepocsort_ref,cbiou_ref,strongsort_ref,hybridsort_ref}, detection bounding boxes of the same objects are joined over the frames and form so called tracklets. In the current frame, new detections are associated with the existing tracklets from the previous frames. This process uses association cues such as object position, motion and displacement information to build a cost matrix reflecting a cost of potential match for each tracklet-detection pair. The association problem is then considered as the bipartite matching problem and solved with the Hungarian matching algorithm~\cite{hungarianalg_ref} to minimize the overall matching cost. Pairs with costs above the pre-defined matching threshold are excluded. Matched detections extend the existing tracklets.

% Following the practices from our baseline, ByteTrack~\cite{bt_ref}, 
% % In case of ByteTrack~\cite{bt_ref}, which we use as our baseline, 
% new detections are split into two groups with higher and lower detection confidence scores and handled in separate association steps. For its association process, the baseline primarily uses intersection over union (IoU). Tracklet position bounding boxes at the current frame are predicted using Kalman Filter~\cite{kf_ref} and are compared to the detection bounding boxes based on the IoU score. Note that IoU $\in [0, 1]$. Cost matrix entries are then filled with 1-IoU score for each tracklet-detection pair. Besides the association part, tracklet management is carefully performed including initiating new tracklets, updating ongoing tracklets and terminating inactive tracklets. We refer the reader to the baseline paper~\cite{bt_ref} for more details if needed.
% 
% --- OLD (CVPR sub) ---
% In our baseline, ByteTrack~\cite{bt_ref}, new detections are split into high and low confidence groups, handled separately in the association process. The baseline uses intersection over union (IoU) as the primary association metric. Tracklet positions are predicted using a Kalman Filter~\cite{kf_ref} and compared to detection bounding boxes using IoU scores. Cost matrix entries are filled with 1-IoU for each tracklet-detection pair. Further, tracklet management includes initiating, updating, and terminating tracklets. For more details, we refer the reader to the ByteTrack paper~\cite{bt_ref}.
%
% --- NEW ---
In our baseline, ByteTrack~\cite{bt_ref}, new detections are split into high and low confidence groups, handled separately in the association process. The baseline uses intersection over union (IoU) as the primary association metric. Tracklet next state position bounding boxes are predicted using a linear motion model, Kalman Filter~\cite{kf_ref}, and compared to the newly observed detection bounding boxes using IoU scores. Cost matrix entries are filled with 1-IoU for each tracklet-detection pair. Tracklet bounding boxes are updated with the matched detection bounding boxes. Unmatched detections are used to initiate new tracklets, while tracklets unmatched for too long are terminated. For more details, we refer the reader to the baseline paper~\cite{bt_ref}.

% --- OLD (CVPR sub) ---
% In our work, we study a temporally propagated segmentation mask as a powerful association cue for MOT. We extend the baseline algorithm~\cite{bt_ref} and combine the mask information with bounding box information to create our novel \textbf{m}asked-\textbf{c}ued algorithm, which we call McByte. \cref{fig:diagram_tracking_pipeline} shows the overview of our tracking pipeline pointing to where the temporally propagated mask is involved as an association cue. 
% % We note that we consider the problem of MOT, where we associate the bounding boxes of tracklets and detections, and not the problems of Single Object Tracking (SOT), Video Object Segmentation (VOS) or Multi-Object Tracking and Segmentation (MOTS). We measure the performance of the bounding box association and not the performance of the mask, which is already reported in the related works [XMem and Cutie refs]. 
% We note that we consider the problem of MOT, where we perform and evaluate the association between the bounding boxes of detections and tracklets.
%
% --- NEW ---
In our work, we study a temporally propagated segmentation mask as a powerful association cue for MOT. We combine the mask and bounding box information to create our novel \textbf{m}asked-\textbf{c}ued algorithm, which we call McByte. \cref{fig:diagram_tracking_pipeline} shows the overview of our tracking pipeline with inclusion of the propagated mask as an association cue. 
% We note that we consider explicitly the problem of MOT, where we perform and evaluate the association between the bounding boxes of detections and tracklets.

% --- Advised by Seongro to comment it out ---
% In the next sections, we describe the creation and handling of the mask within our MOT tracking algorithm (\cref{sec:method_mask_management}) and our regulated use of the temporally propagated mask as an association cue (\cref{sec:method_mask_use}).

\subsection{Mask creation and handling}
\label{sec:method_mask_management}

We design the following approach of mask handling to synchronise it with the processed tracklets.
% --- OLD (CVPR sub) ---
% During tracking in McByte, each tracklet gets its own mask, which is then temporally propagated across frames to update the mask predictions. At first, we use a pre-trained image segmentation model to create a segmentation mask for each new tracklet. It is performed only for a newly appeared object and to initialize a new mask. 
%
% --- NEW ---
Each tracklet gets its own mask, which is then propagated across frames to keep it up-to-date. We use an image segmentation model~\cite{sam_ref} to create a mask for each new tracklet. It is performed only for a newly appeared object and to initialize a new mask. 

% --- OLD (CVPR sub) ---
% Separately, during the next frames, a pre-trained temporal propagator is used to infer the updated mask positions, while aiming to keep the spatio-temporal consistency of the mask. The propagator predictions are analyzed and used in the tracklet-detection association process, as detailed in \cref{sec:method_mask_use}. We manage the temporally propagated masks in sync with the tracklet management system, creating new masks for new tracklets and removing them when a tracklet is terminated. 
%
% --- NEW ---
Separately, during the next frames, a mask propagator model~\cite{cutie_ref} is used to update the object masks. The masks are used in the tracklet-detection association process, as detailed in \cref{sec:method_mask_use}. We process the propagated masks in sync with the tracklet lifespan, creating new masks for new tracklets and removing them when a tracklet is terminated. 

% We provide more details and a diagram including both mask creation and mask temporal propagation in Appendix E. \textbf{\textcolor{green}{(Also updated)}}
%
% Visual examples of the masks being used within our tracker are shown in \cref{fig:visual_differences_1,fig:visual_differences_2}.

% \begin{figure}
% \centering
% % \includegraphics[height=3.0cm]{images/diagram1_mask_management.drawio.png}
% % \includegraphics[width=12cm]{images/Copy of Copy of diagram1_mask_management.drawio(1).png}
% \includegraphics
% [width=8cm]
% {images/mm1_and_mm2.drawio-7.pdf}
% % \vspace*{0.3cm}
% \caption{Cases showing the differences in $mc$ and $mf$ (\cref{sec:method_mask_use}) values of a mask (in blue) within a bounding box. The most optimal case for the mask to provide a good guidance is the second one from the left, where
% % when the mask is sufficiently big and when it is covered by a bounding box as much as possible, i.e 
% both $mc$ and $mf$ are as close to $1$ as possible.}
% % \vspace*{-0.5cm}
% \label{fig:mm1_mm2}
% \end{figure}

\begin{figure}
\centering
% \begin{tabular}{c|c}
% \includegraphics[height=3.0cm]{images/diagram1_mask_management.drawio.png}
% \includegraphics[width=12cm]{images/Copy of Copy of diagram1_mask_management.drawio(1).png}

% was 14cm

% \multicolumn{1}{m{\tempwidth}{\includegraphics[height=2.5cm]{images/mm1_and_mm2.drawio-7.pdf}} \\
% \includegraphics[width=7cm]{images/mm1_and_mm2.drawio-7.pdf} \\
%[width=7cm]
\includegraphics[width=6cm]{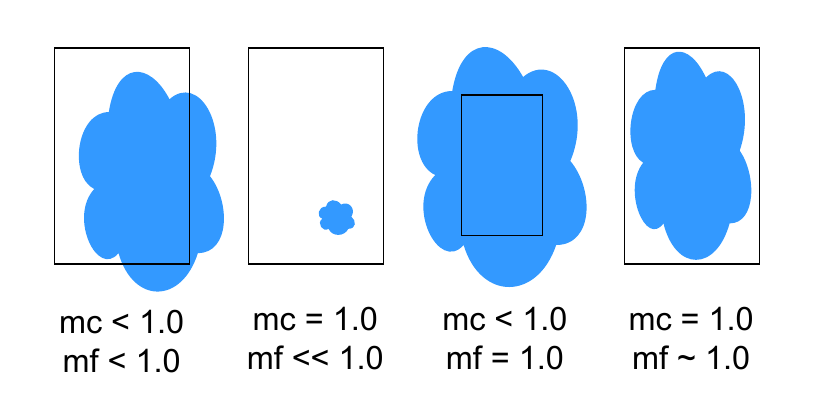} \\
(a) \\
% [width=8cm]
\includegraphics[width=8cm] {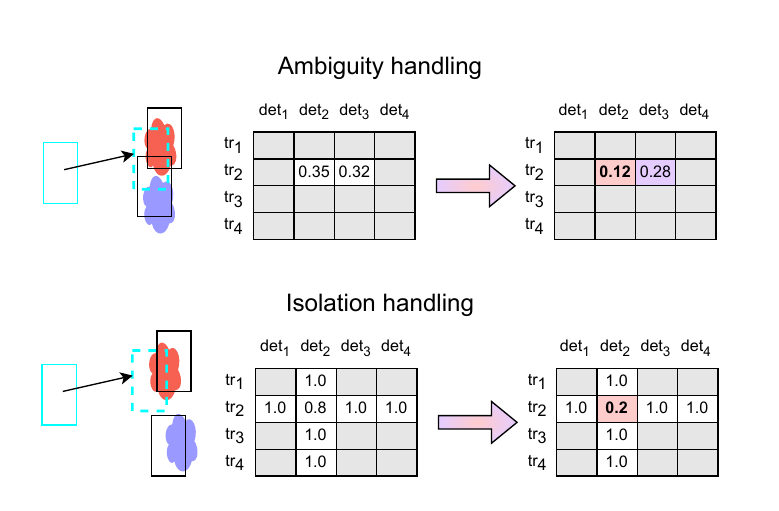} \\ 
(b) \\
% \vspace*{0.3cm}
% \end{tabular}
\caption{(a) Cases showing the differences in $mc$ and $mf$ values of a temporally propagated mask (in blue) within a bounding box. The most optimal case for the mask to provide a good guidance is the last one, where
% when the mask is sufficiently big and when it is covered by a bounding box as much as possible, i.e 
both $mc$ and $mf$ are as close to $1$ as possible. (b) Ambiguity and isolation handling with the mask as a cue. Ambiguity occurs when the IoU-based costs are low and similar for more than one entry in a row (or column) of the cost matrix. Isolation occurs when relevant cost matrix entries contain too high values, not allowing for the association, and at the same time there is no ambiguity.}
\vspace*{-0.3cm}
\label{fig:ambig_isol_handling_and_mm1_mm2}
\end{figure}

\subsection{Regulated use of the mask}
\label{sec:method_mask_use}

% A temporally propagated segmentation mask of a tracked object can be a powerful association cue as long as it is used properly. The mask prediction coming from a temporal mask propagator can sometimes be incorrect as demonstrated on the video segmentation results of the related works~\cite{xmem_ref,cutie_ref}, and thus unreliable. This implies a need for the regulated use of the mask as a cue.
%
% --- OLD (CVPR sub) ---
% A temporally propagated (TP) segmentation mask can be a strong association cue if used correctly. However, mask predictions from the temporal propagator might sometimes be incorrect, as seen in related works~\cite{xmem_ref,cutie_ref} and thus unreliable, making it essential to regulate the mask's use as a cue.
%
% --- NEW ---
% A temporally propagated (TP) segmentation mask can be a strong association cue if used correctly. However,
Mask propagator might sometimes return erroneous masks, as seen in related works~\cite{xmem_ref,cutie_ref}. Therefore, it is essential to regulate the mask use as a guiding association cue.

% During the association process, we use the mask to update the entries of the cost matrix, where the cost values between tracklets and detections are ambiguous. By ambiguity, we mean cases where a tracklet could be potentially matched to more than one detection or where a detection could be matched to more than one tracklet. Such an ambiguity might frequently stem from the matches being based on the IoU score. When the tracked objects are close to each other, their bounding boxes significantly overlap resulting is similar IoU scores and cost values. Therefore, if the IoU is below the matching threshold for more than one tracklet-detection bounding box pair for the same tracklet or detection, we consider it as an ambiguity.
%
% --- OLD (CVPR sub) ---
% In the association process, we update the cost matrix entries using the TP mask, particularly in cases of ambiguity, where a tracklet could match multiple detections or vice versa. Ambiguity often arises from IoU-based matches when tracked objects are close, causing significant overlap in bounding boxes and similar IoU scores. If IoU is below the matching threshold for more than one tracklet-detection pair,
% % of a tracklet or of a detection
% we treat it as ambiguity.
%
% --- NEW ---
In the association process, we update the cost matrix entries using the temporally propagated (TP) mask in two particular, separate cases, for which we refer to as \textit{ambiguity} and \textit{isolation}. In cases of ambiguity, a tracklet could match multiple detections or vice versa. Ambiguity often arises from IoU-based matches when tracked objects are close, causing significant overlap in bounding boxes and thus similar IoU. If IoU-based cost is below the matching threshold 
% (\cref{sec:method_preliminaries}) 
for more than one tracklet-detection pair, we treat it as ambiguity.

In cases of isolation, a detection could potentially match a tracklet without any ambiguity, but their bounding boxes are too far from each other and the overlap is too small. The IoU-based cost is too high (above the matching threshold) to be considered for a match. Isolation might happen when the tracked objects are blurred or when camera movement is abrupt and the next state tracklet bounding boxes cannot be matched with observed detection bounding boxes.

% For each potential ambiguous tracklet-detection match, we apply a strategy consisting of the following conditions.
For each potential ambiguous or isolated tracklet-detection match, we apply a strategy consisting of the following conditions. 
% \begin{enumerate}
% \item 
\textit{(1) We check if the considered tracklet's TP mask is actually visible on the scene.} Subjects can be entirely occluded resulting in no mask prediction at the current frame.
% \item We check if the TP mask prediction is confident enough, i.e. if the average mask probability for the given object from the mask propagator is above the set mask confidence threshold. 
% \item 
\textit{(2) We check if the mask returned by the propagator is confident enough.} We average the confidence probability of all pixels of the mask assigned to a tracklet and check if it is above the set mask confidence threshold.
% \end{enumerate}
Further, we compute two key ratios between the TP mask and a detection bounding box, measured with the number of pixels:
\setlength{\abovedisplayskip}{10pt}
\setlength{\belowdisplayskip}{10pt}
\setlength{\abovedisplayshortskip}{5pt}
\setlength{\belowdisplayshortskip}{5pt}
\begin{itemize}
\item the bounding box coverage of the mask, denoted as $mc$:
\end{itemize}
\begin{equation}
     mc^{i,j} = \frac{|mask(tracklet_{i}) \cap bbox_{j}|}{|mask(tracklet_{i})|} 
  \end{equation}
\begin{itemize}
\item the mask fill ratio of the bounding box, denoted as $mf$:
\end{itemize}
\begin{equation}
    mf^{i,j} = \frac{|mask(tracklet_{i}) \cap bbox_{j}|}{|bbox_{j}|} 
  \end{equation}

where $mask(\cdot)$ denotes the TP mask assigned to the tracklet and $|\cdot|$ denotes the cardinality of the set. Note that all $mc, mf \in [0, 1]$. In \cref{fig:ambig_isol_handling_and_mm1_mm2}(a), we show how $mc$ and $mf$ can vary depending on TP mask and detection bounding box position. 
% We discuss it more in detail in Appendix E.\textbf{\textcolor{red}{[Discuss with Seongro!]}} 
% Going further with the conditions, %:
After that, we consider two more conditions.
% \begin{enumerate}
% \setcounter{enumi}{2}
% \item We check if the mask fill ratio of the bounding box $mf$ occupies a significant portion of the bounding box.
% \item 
\textit{(3) We check if the mask fill ratio of the bounding box $mf$ is sufficiently high.} Very low values can indicate TP masks which are noisy or come from another, wrong tracklet. However, we allow low values reflecting only the visible parts of a tracked object.
\textit{(4) We check if the bounding box coverage of the mask $mc$ is sufficiently high.} Bounding boxes can sometimes be inaccurate, so we allow this value to be slightly below 1.0.
% \end{enumerate}
% Only if all these conditions hold, we update the tracklet-detection association in the cost matrix using the following formula:
Only if all the four conditions hold, we update the corresponding tracklet-detection pair entry in the IoU-based cost matrix:
% using the following formula:
\begin{equation}
  costs^{i,j}=\begin{cases}
    costs^{i,j}_{IoU} - mf^{i,j}, & \text{if cond. \textit{(1)-(4)} satisfied}.%\\
    \\[1ex]
    costs^{i,j}_{IoU}, & \text{otherwise}.
  \end{cases}
  \label{eq:cost_matrix_update}
\end{equation}

% \begin{equation}
%      costs^{i,j} = costs^{i,j}_{IoU} - mf^{i,j}
%      \label{eq:cost_matrix_update_old}
% \end{equation}

% Otherwise, we keep the original 
% % IoU-based 
% cost matrix entry:

% \begin{equation}
%      costs^{i,j} = costs^{i,j}_{IoU}
%      \label{eq:cost_matrix_update_keep_original}
% \end{equation}

where $costs^{i,j}$ denotes the final association cost between the pair of tracklet \textit{i} and detection \textit{j}, and $costs^{i,j}_{IoU}$ denotes the original IoU-based cost for this pair. We apply this process to each entry of the cost matrix, representing each possible tracklet-detection match pair. In \cref{fig:ambig_isol_handling_and_mm1_mm2}(b), we show how TP mask as an association cue influences the cost matrix and guides the association. With this fusion of the available information, we consider both modalities, TP masks and bounding boxes to enhance the association process. In cases of ambiguity or isolation, TP mask assigned to a tracklet can guide the association with a new and suitable detection. The updated cost matrix, enriched by the mask cue, is passed to the Hungarian matching algorithm to find optimal tracklet-detection pairs.

% By following the conditions \textit{1}-\textit{4.}, we ensure that the mask cue is actually controlled and that the cost matrix is updated only if the mask seems reliable. We show the impact of each condition (as well as the lack of any condition) in the ablation study in \cref{sec:exps_abl_stud}, \cref{tab:ablation_mask_constraints_gradual}. More detailed explanations of these conditions are included in the supplementary material, Appendix F. 
%
Following conditions \textit{(1)}-\textit{(4)} ensures that the TP mask cue is controlled, and the cost matrix is updated only when the mask is reliable. 
% The impact of each condition, along with the absence of any, is demonstrated in the ablation study (\cref{sec:exps_abl_stud}, \cref{tab:ablation_mask_constraints_gradual}). 
% Further details on these conditions are included in Appendix E.
%
% Further, the direct use of $mc$ to influence the cost matrix could be misleading, because more than one mask can be completely included within one, the same bounding box, resulting in $mc=1.0$ for all these masks. Hence, we use $mc$ only as a gating condition and use $mf$ to influence the cost matrix.  
% 
Further, using $mc$ directly to influence the cost matrix could be misleading, as multiple TP masks could fully fit within the same bounding box, all resulting in $mc=1.0$. Therefore, we use $mc$ only as a gating condition and  $mf$ to influence the cost matrix.

% Since our baseline is designed to work exactly on the bounding boxes, which is suitable for MOT, we do not change the idea of running the Hungarian matching algorithm over the cost matrix. Instead, we carefully incorporate the mask signal to influence the cost and to improve the association process. In our ablation study in \cref{sec:exps_abl_stud}, we demonstrate that using bounding boxes with mask is actually more beneficial than using the mask on its own. We also show the benefits of adding the mask signal as an association cue over the pure bounding box-based association. 
%
Our baseline is optimized for bounding boxes, so we retain the use of the Hungarian matching algorithm over the cost matrix, but we carefully incorporate the TP mask cue to enhance the association process. 

\subsection{Handling camera motion issues}
\label{sec:method_mcbyte_with_cmc}

% When the video sequence camera is moving, the bounding boxes of tracklets and detections might be less accurate due to the change of object motion and blurry objects. As in our tracklet-detection association process we fuse both the mask information and bounding box information, we add camera motion compensation (CMC) to improve the quality of estimated bounding boxes. We use similar approach as in the literature~\cite{strongsort_ref,deepocsort_ref}. We provide more details about the applied CMC in the supplementary material, Appendix F. The final version of McByte includes CMC and we show its impact in the ablation study in \cref{sec:exps_abl_stud},  \cref{tab:ablation_mask_constraints_gradual}.
%
%
% --- OLD (CVPR sub) ---
% When the camera moves, tracklet and detection bounding boxes may become less accurate due to object motion and blurring. To address this, we also integrate camera motion compensation (CMC) into our process (which fuses temporally propagated mask and bounding box information) to enhance the accuracy of bounding box estimates. This approach follows existing methods~\cite{strongsort_ref,deepocsort_ref}. More details on CMC are provided in Appendix E. The final McByte version includes CMC, and its impact is shown in the ablation study in \cref{sec:exps_abl_stud},  \cref{tab:ablation_mask_constraints_gradual}.
%
% --- NEW ---
% \textbf{\textcolor{red}{[Be careful, so that this one doesn't "steal" the effect/power of the mask, which is our contribution!]}}
When the camera moves, tracklet and detection bounding boxes may become less accurate. To address this, we also integrate camera motion compensation (CMC) into our process to enhance the accuracy of bounding box estimates. Our approach follows the existing methods~\cite{strongsort_ref,deepocsort_ref}. Specifically, we compute a warp (transformation) matrix that accounts for camera movement, based on extracted image features, and apply this matrix to the next state predicted tracklet bounding boxes. This helps adjust for the camera motion, making the tracklet predictions from the Kalman Filter~\cite{kf_ref} and the associations with detections more accurate, improving the overall tracking performance. For key-point extraction, we use the ORB (Oriented FAST and Rotated BRIEF) approach~\cite{orb_ref}.

\section{Experiments and discussion}
\label{sec:experiments_discussion}

\subsection{Implementation details}
\label{sec:exps_impl_details}

For object detections, we follow the baseline~\cite{bt_ref} and use the YOLOX~\cite{yolox_ref} detector pre-trained on the relevant dataset, unless stated otherwise. Detections are split into high and low confidence sets based on a threshold. Unlike the baseline, which adjusts this per sequence, we use a fixed 0.6 threshold across all sequences, matching the baseline’s default when no per-sequence tuning is applied.

For the thresholds of mask confidence, $mc$ and $mf$ (\cref{sec:method_mask_use}) we set the values of 0.6, 0.9 and 0.05 respectively. We set these values considered that they must be high enough depending on their definition. We fix the same values for all sequences and all datasets. Changing these parameters around these values does not affect much performance of McByte making them not sensitive and generic.

% For mask creation, we use SAM~\cite{sam_ref}, ensuring fair comparison with related works, with the ViT\_b model and weights from SAM's original implementation. As a mask temporal propagator, we use Cutie~\cite{cutie_ref} with the weights Cutie base mega. 
% \textbf{\textcolor{red}{[Maybe remove?]}}Note that we use pre-trained SAM only to create and initialize the masks for the newly appeared objects at the scene. 
% For the association cue between the tracklets and detections, we process the propagated mask from the mask temporal propagator (Cutie). 
% All experiments are conducted on an Nvidia A100 GPU with 40 GB of memory.

For mask creation, we use SAM\cite{sam_ref} (ViT\_b model, original weights) for fair comparison with related works. Cutie\cite{cutie_ref} (Cutie base mega weights) is used as the mask temporal propagator.

\vspace*{-0.08cm}

\begin{table}
  \centering
  {\small{
   \scalebox{0.9}{
  \begin{tabular}{@{}lccc@{}}
    \toprule
    Method & HOTA & IDF1 & MOTA \\
    \midrule
    basline~\cite{bt_ref}: no mask & 47.1 & 51.9 & 88.2 \\
    $a1$: either mask or no assoc. & 48.6 & 44.4 & 80.8 \\ 
    $a2$: either mask or IoU for assoc. & 45.3 & 41.5 & 82.2 \\ 
    % ablation 2b \textcolor{magenta}{2a XOR 2b to be removed} & 44.3 & 40.0 & 84.6 \\ 
    $a3$: IoU and mask if ambig. or isol. & 56.6 & 57.0 & 89.5 \\ 
    $a4$: $a3$ + mask confidence & 57.3 & 57.7 & 89.6 \\ 
    $a5$: $a4$ + $mf$ & 58.8 & 60.1 & 89.6 \\ 
    $a6$: $a5$ + $mc$ & 62.1 & 63.4 & 89.7 \\
    % ablation 3e & 62.1 & 63.4 & 89.7 \\ 
    McByte: $a6$ + cmc & \textbf{62.3} & \textbf{64.0} & \textbf{89.8} \\
    \bottomrule
  \end{tabular}
  }
  }}
  \caption{Ablation study on DanceTrack~\cite{dancetrack_ref} validation set listing the effects of the imposed constraints on using the temporally propagated mask as an association cue.}
  \label{tab:ablation_mask_constraints_gradual}
  % \vspace*{-0.3cm}
\end{table}
\begin{table}
  \centering
  {\small{
  \scalebox{0.9}{
  \begin{tabular}{@{}lccc@{}}
    \toprule
    Method & HOTA & IDF1 & MOTA \\
    \midrule 
    Baseline, SportsMOT val                 & 69.0 & 77.9 & 97.5 \\
    McByte, SportsMOT val                   & \textbf{83.9} & \textbf{83.6} & \textbf{98.9} \\
    \midrule
    % \multicolumn{4}{c}{DanceTrack val} \\
    % \midrule
    Baseline, DanceTrack val                & 47.1 & 51.9 & 88.2 \\
    % baseline + M (=abl.3d)  & 62.1 & 63.4 & 89.7 \\ 
    % baseline + LC           & 54.4 & 55.9 & 59.7 \\ 
    McByte, DanceTrack val                  & \textbf{62.3} & \textbf{64.0} & \textbf{89.8} \\ 
    % \textcolor{gray}{baseline + C}            & \textcolor{gray}{53.7} & \textcolor{gray}{54.3} & \textcolor{gray}{89.6} \\ 
    % \textcolor{gray}{baseline + C + M}        & \textcolor{gray}{61.0} & \textcolor{gray}{61.9} & \textcolor{gray}{89.8} \\
    \midrule
    % \multicolumn{4}{c}{SoccerNet-tracking 2022 test} \\
    % \midrule
    % baseline                & \textcolor{lightgray}{71.5} & \textcolor{lightgray}{-} & \textcolor{lightgray}{94.6} \\
    % baseline our run        & 72.1 & 75.3 & 94.5 \\
    Baseline, SoccerNet-tracking 2022 test  & 72.1 & 75.3 & 94.5 \\
    % baseline + M (=abl.3d)  & 82.9 & 77.1 & 96.6 \\
    % baseline + C            & 84.8 & 79.7 & 96.9 \\ 
    McByte, SoccerNet-tracking 2022 test    & \textbf{85.0} & \textbf{79.9} & \textbf{96.8} \\
    \midrule
    % \multicolumn{4}{c}{MOT17 val} \\
    % \midrule
    Baseline, MOT17 val                     & 68.4 & 80.2 & 78.2 \\
    % baseline + M (=abl.3d)  & 69.2 & 81.3 & 78.3 \\
    % baseline + C            & 69.7 & 82.2 & 79.0 \\ 
    McByte, MOT17 val                       & \textbf{69.9} & \textbf{82.8} & \textbf{78.5} \\
    % \midrule
    % % \multicolumn{4}{c}{KITTI-tracking test} \\
    % % \midrule
    % Baseline, KITTI-tracking test           & 54.3 & - & 63.7 \\
    % McByte, KITTI-tracking test             & \textbf{57.0} & - & \textbf{68.9} \\
    \bottomrule
  \end{tabular}
  }
  }}
  \caption{Ablation study comparing McByte with the baseline~\cite{bt_ref} on four different datasets. As SoccerNet-tracking 2022~\cite{soccernet-tracking2022_ref} 
  % and KITTI-tracking~\cite{kitti_ref} (pedestrian) do not
  does not contain validation set split, we report the results on the test set. 
  % KITTI evaluation server does not provide IDF1 scores.
  }
  \label{tab:ablation_mask_cmc}
  \vspace*{-0.3cm}
\end{table}

\subsection{Datasets and evaluation metrics}
\label{sec:exps_datasets_metrics}

We evaluate McByte on four person tracking datasets, primarily in sports settings, while also testing on an additional dataset to assess generalizability. Results are presented on SportsMOT\cite{sportsmot_ref}, DanceTrack\cite{dancetrack_ref}, SoccerNet-tracking 2022\cite{soccernet-tracking2022_ref}, and MOT17\cite{mot17_ref}, using dataset-specific detection sources per community standards for fair comparison.

% SportsMOT~\cite{sportsmot_ref} contains sport scene videos in three different categories: basketball, volleyball and football games. Varying views of playing courts are included in indoor and outdoor settings. Fast camera movement and variable-speed motion of the players are present. We use YOLOX pre-trained on SportsMOT for detections.
% --- Chat GPT ---
SportsMOT~\cite{sportsmot_ref} features basketball, volleyball, and football scenes with diverse indoor and outdoor court views. It includes fast camera movement and variable player speeds. We use YOLOX pre-trained on SportsMOT for detections.

% DanceTrack~\cite{dancetrack_ref} features people performing dances with highly non-linear motion and subtle camera movements, while the number of individuals stays mostly constant. We use YOLOX pre-trained on DanceTrack for detections.
% --- Chat GPT ---
DanceTrack~\cite{dancetrack_ref} includes dancers with highly non-linear motion and subtle camera movements, while the number of individuals remains mostly constant. We use YOLOX pre-trained on DanceTrack for detections.

% SoccerNet-tracking 2022~\cite{soccernet-tracking2022_ref} contains soccer match videos, where players move rapidly and look alike in the same team. Camera movement is always present and oracle detections are provided.
%% by the dataset authors. \textbf{\textcolor{red}{Particularly, SoccerNet-tracking 2022 contains three data splits: train, test and challenge. The challenge set was used to evaluate performance of submitted tracking algorithm outputs during a competition in the past specifically on this set and the algorithms were not included as published research works. Since in our work we compare performance with published works (also on all the other four datasets), see ~\cref{sec:exps_sota_tr_by_det}, we report the results of our method on the test set of SoccerNet-tracking 2022.}}   
% --- Chat GPT ---
SoccerNet-tracking 2022~\cite{soccernet-tracking2022_ref} features soccer match videos with fast-moving players who appear similar within teams. Camera movement is constant, and oracle detections are provided.

% MOT17~\cite{mot17_ref} involves tracking people in public spaces under varying conditions, such as lighting, pedestrian density, and camera stability. We use YOLOX pre-trained on MOT17 for detections. We report the results on MOT17 for reference as it is a well-known dataset in the MOT community. 

MOT17~\cite{mot17_ref} involves tracking pedestrians in public spaces under varying lighting, density, and camera stability. We use YOLOX pre-trained on MOT17 and report results as it is a widely used benchmark in the MOT community.

% KITTI-tracking~\cite{kitti_ref} captures 
% % pedestrian and vehicle 
% scenes from 
% % a moving car,
% car view,
% with varied pedestrian density and frequent camera movement. It also considers another class, car, which can help to assess the generalizability of the tracking method. Following community practices~\cite{ocsort_ref,strongsort_ref}, we use PermaTr~\cite{permatr_ref} detections (also for evaluating our baseline).

% We report three standard MOT metrics: HOTA~\cite{hota_ref}, IDF1~\cite{idf1_ref} and MOTA~\cite{mota_ref}, focusing on HOTA and IDF1 for evaluating the tracking performance. IDF1 measures the tracking quality and identity preservation, while HOTA includes association, detection and localization. MOTA primarily measures detection quality of the tracklets and we report it for the result completeness. Higher values in these metrics indicate better performance.
%% Limited number of submissions is allowed on the test set evaluation servers of MOT17~\cite{mot17_ref} and KITTI-tracking~\cite{kitti_ref}.
% As we focus on the problem of MOT in this work, we report the bounding box association performance using the aforementioned metrics.
% --- ChatGPT ---
We report three standard MOT metrics: HOTA\cite{hota_ref}, IDF1\cite{idf1_ref}, and MOTA~\cite{mota_ref}, focusing on HOTA (association, detection, and localization) and IDF1 (tracking quality and identity preservation). MOTA, which primarily evaluates detection quality, is included for completeness. Higher values indicate better performance.

We remark that we do not train object detectors, and for all datasets we use the same detections as baseline and other compared methods.

\begin{figure}
\centering
% \begin{tabular}{cccccc} % one row option, use {figure*}
\begin{tabular}{ccc} % two row option, use {figure}

Input frames  \\
\includegraphics[height=2.5cm]{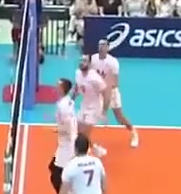}
\includegraphics[height=2.5cm]{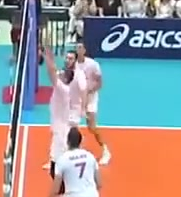}
\includegraphics[height=2.5cm]{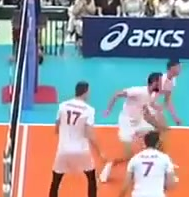}
\\
McByte (ours)  \\
\includegraphics[height=2.5cm]{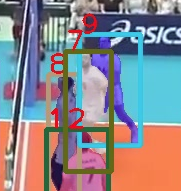}
\includegraphics[height=2.5cm]{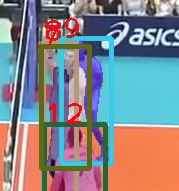}
\includegraphics[height=2.5cm]{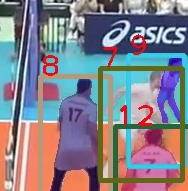}
\\
Baseline \\
\includegraphics[height=2.5cm]{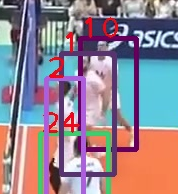}
\includegraphics[height=2.5cm]{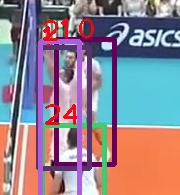}
\includegraphics[height=2.5cm]{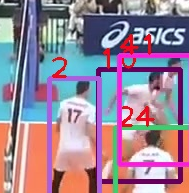}
\\ % for the two row option
% Frame 319 (baseline)&Frame 401 (baseline)&Frame 319 (McByte)&Frame 401 (McByte)
\end{tabular}
\caption{Example comparison with baseline~\cite{bt_ref} in a challenging volleyball setting. McByte can handle the association of  ambiguous sets of bounding boxes. Players IDs 7 and 8 are well maintained despite the temporary jam. ID 9 is also well kept. In case of baseline, the corresponding player IDs are not maintained: After the jam, the player with ID 1 changes their ID to 10 and the player who previously had ID 10, now gets a new ID 41. ID 1 is lost.}
\vspace*{-0.3cm}
\label{fig:visual_differences_volleyball}
\end{figure}

%%% --- HORIZONTAL VERSION ---
% \begin{figure*}
% \centering
% \begin{tabular}{cccc}
% % originally it was: height=4.0cm
% \includegraphics[height=3.5cm]{images/bytetrack_000367.jpg}&
% \includegraphics[height=3.5cm]{images/mcbyte_extra_1_000367.jpg}
% \\
% Baseline & McByte
% \end{tabular}
% \caption{
% Example comparison with baseline~\cite{bt_ref} in a challenging football setting. McByte can maintain the tracklets of the blurry players caused by the abrupt camera movement (pointed by yellow arrows).
% }
% % \vspace*{-0.3cm}
% \label{fig:blurry_football_comparison_new}
% \end{figure*}

%%% --- VERTICAL VERSION ---
\begin{figure}
\centering
\begin{tabular}{cccc}
Baseline\\
% originally it was: height=4.0cm
\includegraphics[height=3.5cm]{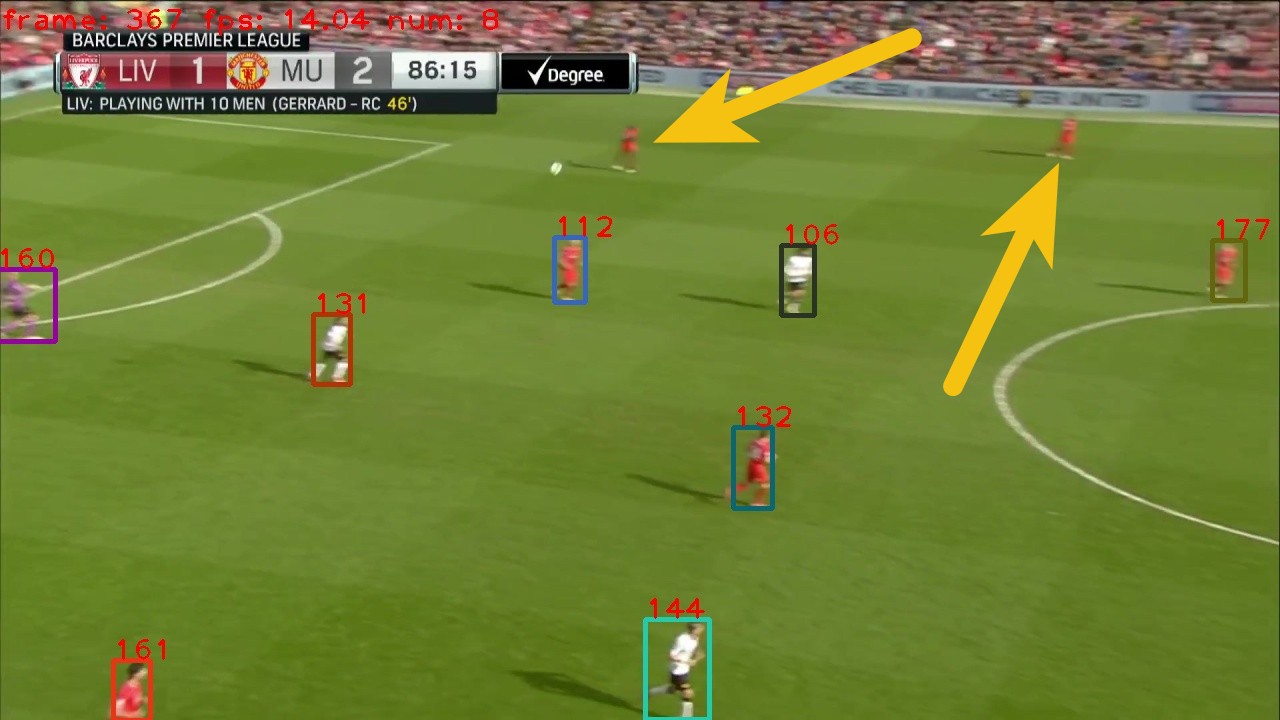}\\
McByte\\
\includegraphics[height=3.5cm]{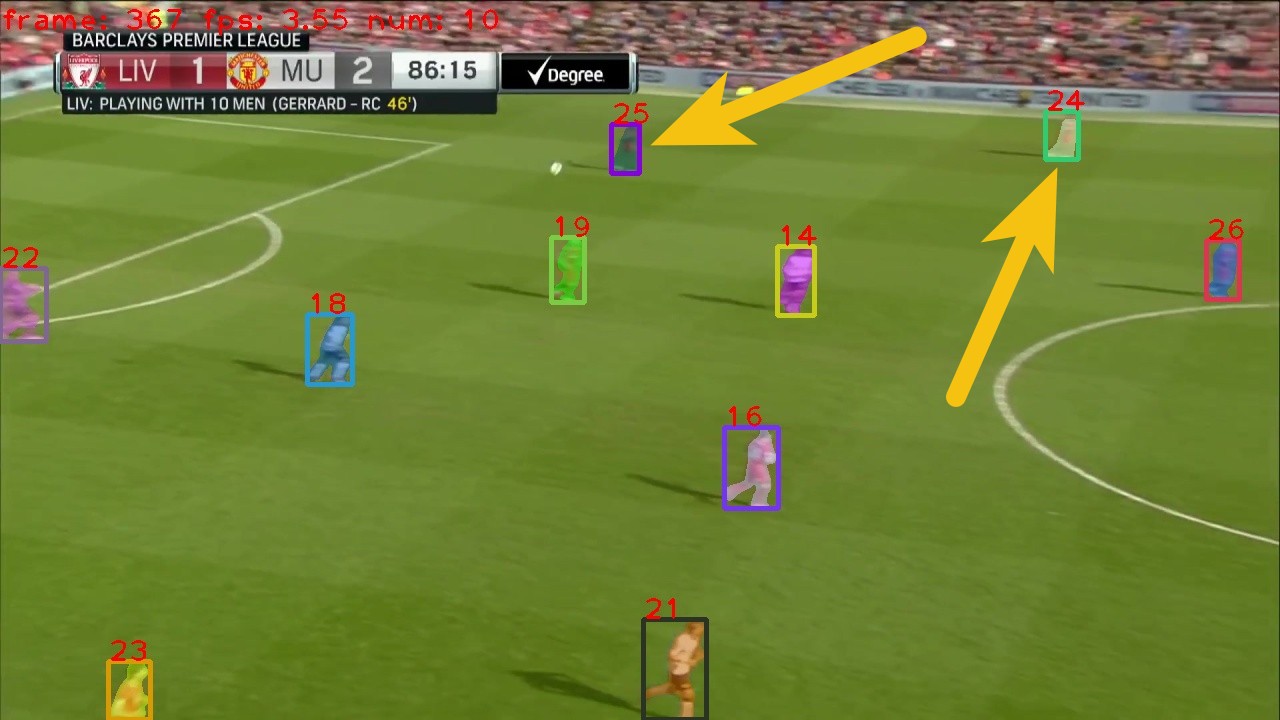}
\end{tabular}
\caption{
Example comparison with baseline~\cite{bt_ref} in a challenging football setting. McByte can maintain the tracklets of the blurry players (pointed by yellow arrows) caused by the abrupt camera movement.
}
\vspace*{-0.5cm}
\label{fig:blurry_football_comparison_new}
\end{figure}

\subsection{Ablation studies}
\label{sec:exps_abl_stud}

We perform an ablation study to demonstrate the impact of incorporating the temporally propagated (TP) mask as an association cue along with the conditions discussed in \cref{sec:method_mask_use}. We evaluate the following variants: 
\begin{itemize}
    \item $a1$: Uses only the TP mask signal for association if the mask is visible for the given tracklet, without ambiguity/isolation checks. The value of $1-mf^{i,j}$ is directly assigned to $costs^{i,j}$ in \cref{eq:cost_matrix_update}. No association occurs if there is no mask.
    \item $a2$: Similar to $a1$, but if the TP mask is unavailable, intersection over union (IoU) scores are used for association.
    \item $a3$: Adds an ambiguity/isolation check. If the TP mask is visible, mask and bounding box information are fused as shown in \cref{eq:cost_matrix_update}. If no mask is available, initial IoU scores are used.
    \item $a4$: Builds on $a3$, incorporating the mask confidence check (condition \textit{(2)}).
    \item $a5$: Extends $a4$ by adding the $mf$ value check from condition \textit{(3)}.
    \item $a6$: Further extends $a5$ with the $mc$ value check from condition \textit{(4)}.
\end{itemize}

The results of each variant are listed in \cref{tab:ablation_mask_constraints_gradual}. In variant $a1$, where only the TP mask signal is used for association, we can see that despite HOTA increase, IDF1 decreases with respect to the baseline. It is caused by the fact that the mask use is uncontrolled and chaotic. With TP mask possibly providing incorrect results, the association cues can be misleading. If we perform the association either based only on TP mask or only on IoU (depending on the availability of the mask), as in variant $a2$, we might face an inconsistency of the cues between tracklets and detections from the same frame and the next frames. This might lead to performance degradation. However, when we use properly both cues fusing the available information (variant $a3$), we can observe significant performance gain. We explain it as the algorithm is initially designed to work on bounding boxes while TP mask is a valuable guiding cue which can improve existing association mechanisms. When the mask signal is the only cue or not properly fused with the IoU cue, a lower performance might be obtained (as in $a1$ and $a2$). 
%
% The results for each variant are presented in \cref{tab:ablation_mask_constraints_gradual}. In variant $a1$, where only the mask signal is used for association, we see an increase in HOTA but a decrease in IDF1 compared to the baseline. This happens because the uncontrolled mask use can lead to incorrect associations, making the process chaotic. In variant $a2$, where the association switches between mask and IoU based on mask availability, the inconsistency between cues can cause mismatches within the same frame or across frames, leading to performance degradation. However, in variant $a3$, where both cues are properly fused, we observe a significant performance improvement. The algorithm, designed primarily for bounding boxes, effectively uses the mask as a guiding signal, improving association without overriding IoU. When the mask is used alone or without proper fusion (as in $a1$ and $a2$), it results in lower overall performance.

\begin{table}
  \centering
  {\small{
  \scalebox{0.9}{
  \begin{tabular}{@{}lccc@{}}
    \toprule
    Method & HOTA & IDF1 & MOTA \\
    \midrule
    ByteTrack~\cite{bt_ref} & 64.1 & 71.4 & 95.9 \\
    MixSort-Byte~\cite{sportsmot_ref} & 65.7 & 74.1 & 96.2 \\
    OC-SORT~\cite{ocsort_ref} & 73.7 & 74.0 & 96.5 \\
    MixSort-OC~\cite{sportsmot_ref} & 74.1 & 74.4 & 96.5 \\
    GeneralTrack~\cite{generaltrack_ref} & 74.1 & 76.4 & 69.8 \\
    DiffMOT~\cite{diffmot_ref} & 76.2 & 76.1 & 97.1 \\
    McByte (ours)                 & \textbf{76.9} & \textbf{77.5} & \textbf{97.2} \\
    %%%
    % \midrule
    % \multicolumn{4}{c}{\textbf{\textcolor{red}{--- POSSIBLY REMOVE THIS PAPER ---}}} \\
    % \multicolumn{4}{c}{\textcolor{magenta}{WACV24 workshop -} \textbf{\textcolor{cyan}{CHECK THE COMMENT} \textcolor{magenta}{in main.tex!}}} \\
    % \midrule
    % \textcolor{magenta}{\st{Deep-EIoU (train)}} & \textcolor{magenta}{\st{74.1}} & \textcolor{magenta}{\st{75.0}} & \textcolor{magenta}{\st{95.1}} \\
    % \textbf{\textcolor{magenta}{Deep-EIoU (train+val)}} & \textbf{\textcolor{magenta}{77.2}} & \textbf{\textcolor{magenta}{79.8}} & \textcolor{magenta}{96.3} \\
    % %%%
    \bottomrule
  \end{tabular}
  }
  }}
  \caption{Comparing McByte with state-of-the-art tracking-by-detection algorithms on SportsMOT test set~\cite{sportsmot_ref}.
  }
  \label{tab:sota_sportsmot_test}
  \vspace*{-0.1cm}
\end{table}

\begin{table}
  \centering
  {\small{
  \scalebox{0.9}{
  \begin{tabular}{@{}lccc@{}}
    \toprule
    Method & HOTA & IDF1 & MOTA \\
    % \midrule
    % \multicolumn{4}{c}{\textcolor{magenta}{Mind extended version! With}} \\
    % \multicolumn{4}{c}{\textcolor{magenta}{Transformers and SUSHI. BMVC Suppl. Mat}} \\
    \midrule
    ByteTrack~\cite{bt_ref}             & 47.7 & 53.9 & 89.6 \\
    MixSort-Byte~\cite{sportsmot_ref} & 46.7 & 53.0 & 85.5 \\
    OC-SORT~\cite{ocsort_ref}           & 55.1 & 54.9 & 92.2 \\
    Deep OC-SORT~\cite{deepocsort_ref}  & 61.3 & 61.5 & 92.3 \\
    % \textcolor{blue}{C-BIoU~\cite{cbiou_ref} as reported (no code)}             & \textcolor{blue}{60.6} & \textcolor{blue}{61.6} & \textcolor{blue}{91.6} \\
    % C-BIoU~\cite{cbiou_ref} * our impl.          & 45.8 & 52.0 & 88.4 \\
    StrongSORT++~\cite{strongsort_ref}  & 55.6 & 55.2 & 91.1 \\
    Hybrid-SORT~\cite{hybridsort_ref}   & 65.7 & 67.4 & 91.8 \\
    GeneralTrack\cite{generaltrack_ref} & 59.2 & 59.7 & 91.8 \\
    DiffMOT~\cite{diffmot_ref} & 63.4 & 64.0 & 92.7 \\
    % Ours (McByte)            & 66.5 & \textbf{68.2} & 92.7 \\
    McByte (ours)                       & \textbf{67.1} & \textbf{68.1} & \textbf{92.9} \\
    \bottomrule
  \end{tabular}
  }
  }}
  \caption{Comparing McByte with state-of-the-art tracking-by-detection algorithms on DanceTrack test set~\cite{dancetrack_ref}.
  % *We reproduce C-BIoU and report the best result obtained.
  }
  \label{tab:sota_dancetrack_test}
\end{table}
\begin{table}
  \centering
  {\small{
  \scalebox{0.9}{
  \begin{tabular}{@{}lccc@{}}
    \toprule
    Method & HOTA & IDF1 & MOTA \\
    \midrule
    % ByteTrack~\cite{bt_ref}   & \textcolor{lightgray}{71.5} & \textcolor{lightgray}{94.6} & \textcolor{lightgray}{-}\\
    ByteTrack~\cite{bt_ref}       & 72.1 & 75.3 & 94.5 \\
    % DeepSORT~\cite{deepsort_ref} & 69.6 & 94.8 & -\\
    OC-SORT~\cite{ocsort_ref}     & 82.0 & 76.3 & \textbf{98.3} \\
    % C-BIoU~\cite{cbiou_ref}    & \textcolor{lightgray}{\textbf{89.2}} & \textcolor{lightgray}{\textbf{86.1}} & \textcolor{lightgray}{\textbf{99.4}} \\
    % \textcolor{blue}{C-BIoU~\cite{cbiou_ref} as reported (no code)}             & \textcolor{blue}{\textbf{\underline{89.2}}} & \textcolor{blue}{\textbf{\underline{86.1}}} & \textcolor{blue}{\textbf{\underline{99.4}}} \\
    % C-BIoU~\cite{cbiou_ref} * our impl.       & 72.7 & 76.4 & 95.4 \\
    % Ours (McByte)             & 82.9 & 77.1 & 96.6 \\
    McByte (ours)                 & \textbf{85.0} & \textbf{79.9} & 96.8 \\
    % \textbf{\textcolor{magenta}{Deep-EIoU (WACV workshop)}} & \textbf{\textcolor{magenta}{85.4}} & \textbf{\textcolor{magenta}{-}} & \textcolor{magenta}{-} \\
    % \multicolumn{4}{c}{\textbf{\textcolor{red}{--- POSSIBLY REMOVE THE PAPER ABOVE ---}}} \\
   
    \bottomrule
  \end{tabular}
  }
  }}
  \caption{Comparing McByte with state-of-the-art tracking-by-detection algorithms on SoccerNet-tracking 2022 test set~\cite{soccernet-tracking2022_ref}.
  % *We reproduce C-BIoU and report the best result obtained.
  }
  \label{tab:sota_soccernet_test}
\end{table}

% \begin{table}
%   \centering
%   {\small{
%   \colorbox{lightgray}{%
%   \begin{tabular}{@{}lccc@{}}
%     \toprule
%     Tracking challenge & HOTA & DetA & AssA \\
%     \midrule
%     SoccerNet 2022, challenge split              & 86.0 & 97.2 & 76.0\\
%     SoccerNet 2023, challenge split              & 64.3 & 67.4 & 61.4\\
%     \bottomrule
%   \end{tabular}
%   }%
%   }}
%   \caption{\textcolor{orange}{\textbf{[Currently not referred in th text] }}\\ \textcolor{red}{\textbf{[GENERALLY WORSE THAN SEVERAL BEST ONES]} McByte on SoccerNet-tracking 2022 (oracle dets) and 2023 (same YOLOX as for SportsMOT challenge splits. Tables with other methods available here: https://www.soccer-net.org/tasks/tracking}}
%   % \vspace*{-0.5cm}
%   \label{tab:mcbyte_soccernet_challenges}
% \end{table}

\begin{table}
  \centering
  {\small{
  \scalebox{0.9}{
  \begin{tabular}{@{}lccc@{}}
    \toprule
    Method & HOTA & IDF1 & MOTA \\
    % \midrule
    % \multicolumn{4}{c}{\textcolor{magenta}{Mind extended version! With}} \\
    % \multicolumn{4}{c}{\textcolor{magenta}{Transformers and SUSHI. BMVC Suppl. Mat}} \\
    \midrule
    \multicolumn{4}{c}{With parameter tuning per sequence} \\
    \midrule
    \textcolor{gray}{ByteTrack~\cite{bt_ref}}             & \textcolor{gray}{63.1} & \textcolor{gray}{77.3} & \textcolor{gray}{80.3} \\
    \textcolor{gray}{MixSort-Byte~\cite{sportsmot_ref}}             & \textcolor{gray}{64.0} & \textcolor{gray}{78.7} & \textcolor{gray}{79.3} \\
    % BoT-SORT~\cite{botsort_ref}           & 65.0 & 80.2 & 80.5 \\
    \textcolor{gray}{StrongSORT++~\cite{strongsort_ref}}  & \textcolor{gray}{64.4} & \textcolor{gray}{79.5} & \textcolor{gray}{79.6} \\
    % \textcolor{violet}{C-BIoU~\cite{cbiou_ref}}          & \textcolor{violet}{64.1} & \textcolor{violet}{79.7} & \textcolor{violet}{81.1} \\
    % C-BIoU~\cite{cbiou_ref} *           & TO & BE & RUN \\
    % ImprAsso~\cite{imprasso_ref}        & \textbf{66.4} & \textbf{82.1} & \textbf{82.2} \\
    \textcolor{gray}{OC-SORT~\cite{ocsort_ref}}           & \textcolor{gray}{63.2} & \textcolor{gray}{77.5} & \textcolor{gray}{78.0} \\
    \textcolor{gray}{MixSort-OC~\cite{sportsmot_ref}}           & \textcolor{gray}{63.4} & \textcolor{gray}{77.8} & \textcolor{gray}{78.9} \\
    \textcolor{gray}{Deep OC-SORT~\cite{deepocsort_ref}}  & \textcolor{gray}{64.9} & \textcolor{gray}{80.6} & \textcolor{gray}{79.4} \\
    \textcolor{gray}{Hybrid-SORT~\cite{hybridsort_ref}}   & \textcolor{gray}{64.0} & \textcolor{gray}{78.7} & \textcolor{gray}{79.9} \\
    % \midrule
    % \multicolumn{4}{c}{Without parameter tuning per sequence} \\
    % \midrule
    % \multicolumn{4}{c}{\textbf{\textcolor{magenta}{ByteTrack: with our without interpolation?}}} \\
    % \multicolumn{4}{c}{\textcolor{magenta}{Ours: WITH INTERPOLATION}} \\
    % \multicolumn{4}{c}{\textcolor{magenta}{ByteTrack with no param tuning reported}} \\
    % \multicolumn{4}{c}{\textcolor{magenta}{by GHOST and SUSHI: WITHOUT INTERPOLATION}} \\
    \midrule
    % \multicolumn{4}{c}{No implementation provided} \\
    \multicolumn{4}{c}{Without parameter tuning per sequence} \\
    \midrule
    ByteTrack~\cite{sushi_ref} & 62.8 & 77.1 & 78.9 \\
    % \textcolor{blue}{C-BIoU~\cite{cbiou_ref} as reported (no code)}             & \textcolor{blue}{64.1} & \textcolor{blue}{\textbf{\underline{79.7}}} & \textcolor{blue}{\textbf{\underline{81.1}}} \\
    % C-BIoU~\cite{cbiou_ref} * our impl.          & 62.4 & 77.1 & 79.5 \\
    % \midrule
    % \multicolumn{4}{c}{Without parameter tuning per sequence} \\
    % \midrule
    % \textcolor{red}{[REMOVE IT]} \textcolor{magenta}{ByteTrack (with interp.)} & 63.2 & 77.5 & 79.7 \\
    % Ours (McByte)                    & 63.1 & 78.0 & 79.7 \\
    GeneralTrack ~\cite{generaltrack_ref} & 64.0 & 78.3 & \textbf{80.6} \\
    DiffMOT ~\cite{diffmot_ref} & \textbf{64.2} & 79.3 & 79.8 \\
    McByte (ours)                      & \textbf{64.2} & \textbf{79.4} & 80.2 \\
    \bottomrule
  \end{tabular}
  }
  }}
  \caption{Comparing McByte with state-of-the-art tracking-by-detection algorithms on MOT17 test set~\cite{mot17_ref}.
  % *We reproduce C-BIoU and report the best result obtained.
  }
  \label{tab:sota_mot17_test}
 \vspace*{-0.3cm}
\end{table}

Adding the conditional check based on TP mask confidence (variant $a4$) further improves the performance, because sometimes mask might be uncertain or incorrect providing misleading association guidance. Adding the minimal $mf$ value check (variant $a5$) also provides performance gain, because this check filters out the tracklet TP masks which could be considered as a noise or tiny parts of people almost entirely occluded. Another performance gain can be observed with the minimal $mc$ value check (variant $a6$). This check determines if the TP mask of the tracked person is actually within the bounding box and not too much outside it. Since the detection box might not be perfect, we allow small parts of the tracklet mask to be outside the bounding box, but its major part must be within the bounding box, so that the TP mask can be used for guiding the association between the considered tracklet-detection pair. As it is shown, it further helps. Finally, we add the camera motion compensation (CMC), denoted as McByte in \cref{tab:ablation_mask_constraints_gradual}, which also provides some performance gain. 
We compare our McByte tracking algorithm to the baseline~\cite{bt_ref} across the four datasets, as shown in \cref{tab:ablation_mask_cmc}. Performance gain is significant as a whole since improvement is always present, in all cases examined. DanceTrack~\cite{dancetrack_ref} features mostly continuous non-linear motion and many substantial occlussions, where the TP mask signal is particularly helpful in tracking the subjects, whereas IoU might struggle.
SportsMOT~\cite{sportsmot_ref} and SoccerNet-tracking 2022~\cite{soccernet-tracking2022_ref} feature similar outfits among the players, more abrupt motion, blurry objects, e.g. due to fast camera movement after the ball, and occlusions caused by the nature of the team sports. TP mask can handle these situations very well as shown in \cref{fig:visual_differences_volleyball,fig:blurry_football_comparison_new}, and further improves the performance. Full frame images of \cref{fig:visual_differences_volleyball} and more visual examples are available in \cref{sec:app_f_more_visuals} due to the space limits.
Although MOT17~\cite{mot17_ref} is not sports-specific, it is widely used in the MOT community, so we evaluate McByte on it. As shown in \cref{tab:ablation_mask_cmc}, McByte achieves competitive performance and consistently improves tracking scores, demonstrating its robustness and general applicability.

\begin{table*}
  \centering
  {\small{
  \scalebox{0.75}{ % previously used scale: 0.915
  \begin{tabular}{@{}l|ccc|ccc|ccc|ccc@{}}%|cc@{}}
    \toprule
     & \multicolumn{3}{c|}{SoccerNet-tracking 2022} & \multicolumn{3}{c|}{MOT17} & \multicolumn{3}{c|}{DanceTrack} & \multicolumn{3}{c}{SportsMOT} \\ %& \multicolumn{2}{c}{KITTI-tracking} \\
    \midrule
    Method & HOTA & IDF1 & MOTA & HOTA & IDF1 & MOTA & HOTA & IDF1 & MOTA & HOTA & IDF1 & MOTA \\ %& HOTA & MOTA\\
    \midrule 
    McByte (ours)         & 85.0 & 79.9 & 96.8 & 64.2 & 79.4 & 80.2 & 67.1 & 68.1 & 92.9 & 76.9 & 77.5 & 97.2 \\ %& 57.0 & 68.9 \\
    OC-SORT~\cite{ocsort_ref}        & 82.0 & 76.3 & 98.3 & 63.2 & 77.5 & 78.0 & 55.1 & 54.9 & 92.9 & 73.7 & 74.0 & 96.5 \\ %& 54.7 & 65.1 \\  
    C-BIoU~\cite{cbiou_ref}        & 89.2 & 86.1 & 99.4 & 64.1 & 79.7 & 81.1 & 60.6 & 61.6 & 91.6 & - & - & - \\ %& - & - \\
    C-BIoU impl.   & 72.2 & 76.4 & 95.4 & 62.4 & 77.1 & 79.5 & 45.8 & 52.0 & 88.4 & - & - & - \\ %& - & - \\
    \bottomrule
  \end{tabular}
  } % scalebox
  }}
  \caption{Comparison with the tracking-by-detection methods which do not require training and use the same detections (\cref{sec:exps_no_train_methods}). Comparing with C-BIoU is not direct as it involves manual tuning of important hyper-parameters. Since C-BIoU does not provide the code, we implement it following all the information in the paper (\textit{"C-BIoU impl."}).}

  \label{tab:no_train_methods_horizontal}
  \vspace*{-0.3cm}
\end{table*}

\subsection{Comparison with state of the art tracking-by-detection methods}
\label{sec:exps_sota_tr_by_det}

% We compare our McByte with state of the art tracking-by-detection algorithms on the test sets of all four datasets mentioned. The results are listed in \cref{tab:sota_dancetrack_test,tab:sota_mot17_test,tab:sota_soccernet_test,tab:sota_kitti_test}. Our proposed McByte reaches the highest HOTA and IDF1 scores on DanceTrack~\cite{dancetrack_ref} (\cref{tab:sota_dancetrack_test}), on SoccerNet-tracking 2022~\cite{soccernet-tracking2022_ref} (\cref{tab:sota_soccernet_test}) and the highest HOTA scores on KITTI-tracking~\cite{kitti_ref} (\cref{tab:sota_kitti_test}). 
%
We compare McByte with state-of-the-art tracking-by-detection algorithms across the test sets of the four diverse datasets. 
% \textbf{\textcolor{magenta}{[Here or at the end] As described/explained in \cref{sec:related_work__tranformers}, we provide more extensive comparison in Appendix B.}}
% \textbf{\textcolor{red}{[Keep or skip? (the part that follows)]}}We compare our method to general published MOT works evaluated on several datasets rather than methods specifically designed for one dataset. In case of SoccerNet-tracking 2022~\cite{soccernet-tracking2022_ref}, we compare our method to the published works on the test set.
The results in \cref{tab:sota_sportsmot_test,tab:sota_dancetrack_test,tab:sota_soccernet_test,tab:sota_mot17_test} show that McByte outperforms the other methods on HOTA, IDF1 and MOTA on SportsMOT~\cite{sportsmot_ref} (\cref{tab:sota_sportsmot_test}) and DanceTrack~\cite{dancetrack_ref} (\cref{tab:sota_dancetrack_test}). In case of SoccerNet-tracking 2022~\cite{soccernet-tracking2022_ref} (\cref{tab:sota_soccernet_test}) and MOT17~\cite{mot17_ref} (\cref{tab:sota_mot17_test}), McByte outperforms the other methods on HOTA and IDF1, and achieves the second best MOTA score. 
% All the listed methods use the same detections per dataset in \cref{tab:sota_sportsmot_test,tab:sota_dancetrack_test,tab:sota_soccernet_test,tab:sota_mot17_test,tab:sota_kitti_test}. 
We note that MOTA metrics mostly reflect the detection quality of the tracklets.
% As oracle detections, meaning perfect detections, are provided by the dataset authors~\cite{soccernet-tracking2022_ref}, we argue that MOTA is not very descriptive in terms of the actual tracking performance.
Our proposed method works very well on datasets including settings with different sports, i.e. basketball, volleyball, football and dance. Further, it also achieves satisfactory results on another dataset involving pedestrians with less dynamic and more linear movement which, we believe, demonstrates its generalizability and wide applicability. While we do not perform any costly training or tuning on the evaluated datasets, our method still reaches surpassing results on HOTA and IDF1 and competitve results on MOTA.

On MOT17~\cite{mot17_ref}, unlike the baseline~\cite{bt_ref} and its extensions~\cite{strongsort_ref, ocsort_ref, deepocsort_ref, hybridsort_ref}, we do not tune parameters per sequence, aiming for a generalizable tracker. Among non-tuned trackers, McByte achieves the best scores. We also include the result of ByteTrack~\cite{bt_ref} not being tuned per sequence as reported in \cite{sushi_ref} (\textit{"ByteTrack~\cite{sushi_ref}"}).  Compared to tuned methods, McByte improves over the baseline~\cite{bt_ref} and remains competitive with other approaches.

% --- NEW, extra added ---

The other types of tracking methods, such as transformer-based~\cite{motip_ref,motrv2_ref,memotr_ref, motr_ref}, global optimization~\cite{sushi_ref} and joint detection and tracking~\cite{fairmot_ref,relationtrack_ref} require a lot of training data and might use other detections, which is not the focus of this work. For reference, we list their results together with other approaches and ours in \cref{sec:sota_transformers_sushi}.

% We also remark that C-BIoU~\cite{cbiou_ref} method is placed with asterix in \cref{tab:sota_dancetrack_test,tab:sota_mot17_test,tab:sota_soccernet_test}. For this algorithm, no implementation is published and not all the necessary details for reproduction are provided. We reproduce the method based on all the description provided and report the best results we have managed to obtain. We describe it more in detail in the supplementary material, Appendix C.
%
% %
% {\huge\textbf{\textcolor{red}{[To be discussed! And decided...]}}}Additionally, C-BIoU~\cite{cbiou_ref} is marked with an asterisk in \cref{tab:sota_dancetrack_test,tab:sota_mot17_test,tab:sota_soccernet_test}. As no implementation is published, and not all necessary details for reproduction are provided, we reproduce the method based on the available descriptions and report the best results we have obtained. We describe it more in detail in Appendix C.

\subsection{Comparison with non-trainable methods}
\label{sec:exps_no_train_methods}

In \cref{tab:no_train_methods_horizontal}, we compare McByte with other non-trainable tracking-by-detection algorithms using the same detections. McByte, OC-SORT~\cite{ocsort_ref}, and C-BIoU~\cite{cbiou_ref} achieve strong performance. McByte outperforms OC-SORT on all datasets, though OC-SORT achieves higher MOTA (reflecting the tracklet detection accuracy) on SoccerNet-tracking 2022, where oracle detections are provided.

% In case of C-BIoU~\cite{cbiou_ref}, the comparison is not direct as C-BIoU involves human-dependent tuning and setting per each dataset the buffer scales, critical hyper-parameters as mentioned in the paper. Their values are provided for DanceTrack in the main work~\cite{cbiou_ref}. In a separate technical report~\cite{cbiou_tech_report_ref} of the C-BIoU algorithm, different values are listed for the buffer scales and for the matching threshold for the SoccerNet-tracking~\cite{soccernet-tracking2022_ref} dataset. In our case, we always use the exact same value of each hyper-parameter for all datasets and all sequences. Unfortunately, C-BIoU code is not provided. However, the the authors kindly list some of the technical details and parameters, based on which we implement the C-BIoU algorithm. We list the results on the originally reported datasets in \cref{tab:no_train_methods_horizontal} (\textit{"C-BIoU impl."}) without changing the parameters per dataset. Differences in the performance reflect the sensitivity of the hyper-parameters. 
% --- Chat GPT ---
C-BIoU~\cite{cbiou_ref} is not directly comparable, as it requires human-dependent tuning of buffer scales, critical hyper-parameters for each dataset. Values are provided for DanceTrack~\cite{dancetrack_ref}, while a separate technical report~\cite{cbiou_tech_report_ref} lists different settings for SoccerNet-tracking~\cite{soccernet-tracking2022_ref}. Unfortunately, C-BIoU’s code is not provided. However, the authors kindly share technical details, allowing us to implement it with fixed hyper-parameters across all datasets ("C-BIoU impl."). Differences in the performance reflect the sensitivity to the hyper-parameter tuning.

\subsection{Comparison with other methods using mask}
\label{sec:exps_other_mask_methods}

We evaluate mask-based tracking methods DEVA~\cite{deva_ref}, Grounded SAM 2~\cite{sam2_ref,gr_dino_ref}, and MASA~\cite{masa_ref} on the SportsMOT validation set. Each method is tested with its original settings and with YOLOX~\cite{yolox_ref} trained on SportsMOT, as used in our baseline~\cite{bt_ref} and McByte for fair comparison. Results in \cref{tab:deva_grsam2_masa_sportsmotval} show that McByte outperforms all other mask-based methods.

% DEVA~\cite{deva_ref} lacks a tracklet management system, resulting in negative MOTA scores, which occur when errors exceed the number of objects~\cite{motchallenge_paper_ref}. Adding YOLOX detections relatively improves the performance by limiting the considered objects, but the performance remains unsatisfactory. Grounded SAM 2~\cite{sam2_ref,gr_dino_ref} works with segment-based tracking, but merging segments into full tracklets can be unreliable. MASA~\cite{masa_ref} cannot handle longer than few-frame occlusions and might miss some detections
% %and both, misses detections and processes redundant ones (e.g. from the audience) in its original setup, 
% yielding low MOT performance. 
% --- Chat GPT ---
DEVA\cite{deva_ref} lacks a tracklet management system, leading to negative MOTA scores when errors exceed object counts\cite{motchallenge_paper_ref}. Adding YOLOX improves performance by refining object selection but remains limited. Grounded SAM 2\cite{sam2_ref,gr_dino_ref} uses segment-based tracking, though merging segments into tracklets can be inconsistent. MASA\cite{masa_ref} has difficulty handling longer occlusions, occasionally missing detections, which affects MOT performance.

These results highlight that existing mask-based methods are unsuitable for MOT, particularly in sports. In contrast, McByte effectively combines temporally propagated mask-based association with bounding box processing and tracklet management, making it better suited for MOT tasks in sports and beyond.

More results and details on DEVA, Grounded SAM 2, and MASA experiments, also on other datasets (DanceTrack~\cite{dancetrack_ref}, MOT17~\cite{mot17_ref}), are available in \cref{sec:mask_based_systems_more_exps}.

\begin{table}
  \centering
  {\small{
  \scalebox{0.9}{
  \begin{tabular}{@{}lccc@{}}
    \toprule
    Method & HOTA & IDF1 & MOTA \\
    \midrule
    DEVA, original settings        & 39.3 & 37.3 & -109.6 \\
    DEVA, with YOLOX               & 42.4 & 42.1 & -57.0 \\
    \midrule
    Grounded SAM 2, original settings & 45.9 & 44.7 & -13.9 \\
    % step=20
    Grounded SAM 2, with YOLOX     & 66.1 & 70.2 & 91.4 \\
    % step=20
    \midrule
    MASA, original settings        & 39.4 & 35.7 & -27.2 \\
    MASA, with YOLOX               & 73.6 & 71.2 & 97.0 \\
    \midrule
    McByte (ours)                  & \textbf{83.9} & \textbf{83.6} & \textbf{98.9} \\
    \bottomrule
  \end{tabular}
  }
  }}
  \caption{Comparison with the other tracking methods using segmentation mask: DEVA~\cite{deva_ref}, Grounded SAM 2~\cite{sam_ref,gr_dino_ref} and MASA~\cite{masa_ref} on SportsMOT validation set~\cite{sportsmot_ref}. We compare the variants with original settings and with the same object detector (YOLOX) as in McByte.}
  \label{tab:deva_grsam2_masa_sportsmotval}
  \vspace*{-0.30cm}
\end{table}

% \begin{table}
%   \centering
%   {\small{
%   \colorbox{gray}{%
%   \begin{tabular}{@{}lccc@{}}
%     \toprule
%     Method & HOTA & IDF1 & MOTA \\
%     \midrule
%     DEVA, original settings        & 31.8 & 31.3 & -89.4 \\
%     DEVA, with YOLOX               & 24.7 & 20.4 & -239.7 \\
%     \midrule
%     Grounded SAM 2, original settings & 43.4 & 47.6 & 18.4 \\
%     % step=20
%     Grounded SAM 2, with YOLOX     & 47.5 & 54.1 & 43.0 \\
%     % step=20
%     \midrule
%     MASA, original settings        & 45.5 & 53.6 & 36.9 \\
%     MASA, with YOLOX               & 63.5 & 73.6 & 74.0 \\
%     \midrule
%     McByte (ours)                  & \textbf{69.9} & \textbf{82.8} & \textbf{78.5} \\
%     \bottomrule
%   \end{tabular}
%   }%
%   }}
%   \caption{\textcolor{orange}{\textbf{[Currently not referred in th text] }}\\Comparison with the other tracking methods using segmentation mask: DEVA~\cite{deva_ref}, Grounded SAM 2~\cite{sam_ref,gr_dino_ref} and MASA~\cite{masa_ref} on MOT17 validation set~\cite{mot17_ref}. }
%   \label{tab:deva_grsam2_masa_mot17val}
%   % \vspace*{-0.30cm}
% \end{table}

\section{Conclusion}
\label{sec:conclusion}

We introduce a temporally propagated segmentation mask as an association cue for MOT, focusing on sports tracking. Our approach incorporates mask propagation into tracking-by-detection by fusing mask and bounding box information while following our practical policies to enhance the performance. McByte requires no training or tuning, relying only on pre-trained models and object detectors for fair comparison. Results across four datasets highlight its effectiveness in sports and its generalizability to other scenarios.

\section*{Acknowledgement}
This work has been supported by the French government, through the 3IA Cote d’Azur Investments in the project managed by the National Research Agency (ANR) with the reference number ANR-23-IACL-0001. 

This work was granted access to the HPC resources of IDRIS under the allocation 2024-AD011014370 made by GENCI. 

% \newpage
% \pagebreak

%%%%%%%%% REFERENCES
{
    \small
    \bibliographystyle{ieeenat_fullname}
    \bibliography{main}

\begin{thebibliography}{44}
\providecommand{\natexlab}[1]{#1}
\providecommand{\url}[1]{\texttt{#1}}
\expandafter\ifx\csname urlstyle\endcsname\relax
  \providecommand{\doi}[1]{doi: #1}\else
  \providecommand{\doi}{doi: \begingroup \urlstyle{rm}\Url}\fi

\bibitem[Atkinson and Shiffrin(1968)]{atk_shf_mem_model_ref}
R.C. Atkinson and R.M. Shiffrin.
\newblock Human memory: A proposed system and its control processes.
\newblock pages 89--195. Academic Press, 1968.

\bibitem[Bernardin and Stiefelhagen(2008)]{mota_ref}
Keni Bernardin and Rainer Stiefelhagen.
\newblock Evaluating multiple object tracking performance: The clear mot metrics.
\newblock \emph{EURASIP Journal on Image and Video Processing}, 2008, 2008.

\bibitem[Bhattacharjee et~al.(2023)Bhattacharjee, Süsstrunk, and Salzmann]{adapters_cv_ref}
Deblina Bhattacharjee, Sabine Süsstrunk, and Mathieu Salzmann.
\newblock Vision transformer adapters for generalizable multitask learning, 2023.

\bibitem[Cao et~al.(2023)Cao, Pang, Weng, Khirodkar, and Kitani]{ocsort_ref}
Jinkun Cao, Jiangmiao Pang, Xinshuo Weng, Rawal Khirodkar, and Kris Kitani.
\newblock Observation-centric sort: Rethinking sort for robust multi-object tracking.
\newblock In \emph{Proceedings of the IEEE/CVF Conference on Computer Vision and Pattern Recognition (CVPR)}, pages 9686--9696, 2023.

\bibitem[Cetintas et~al.(2023)Cetintas, Bras\'o, and Leal-Taix\'e]{sushi_ref}
Orcun Cetintas, Guillem Bras\'o, and Laura Leal-Taix\'e.
\newblock Unifying short and long-term tracking with graph hierarchies.
\newblock In \emph{Proceedings of the IEEE/CVF Conference on Computer Vision and Pattern Recognition (CVPR)}, pages 22877--22887, 2023.

\bibitem[Cheng and Schwing(2022)]{xmem_ref}
Ho~Kei Cheng and Alexander~G. Schwing.
\newblock {XMem}: Long-term video object segmentation with an atkinson-shiffrin memory model.
\newblock In \emph{Proceedings of the European Conference on Computer Vision (ECCV)}, 2022.

\bibitem[Cheng et~al.(2023)Cheng, Oh, Price, Schwing, and Lee]{deva_ref}
Ho~Kei Cheng, Seoung~Wug Oh, Brian Price, Alexander Schwing, and Joon-Young Lee.
\newblock Tracking anything with decoupled video segmentation.
\newblock In \emph{Proceedings of the IEEE/CVF International Conference on Computer Vision (ICCV)}, 2023.

\bibitem[Cheng et~al.(2024)Cheng, Oh, Price, Lee, and Schwing]{cutie_ref}
Ho~Kei Cheng, Seoung~Wug Oh, Brian Price, Joon-Young Lee, and Alexander Schwing.
\newblock Putting the object back into video object segmentation.
\newblock In \emph{Proceedings of the IEEE/CVF Conference on Computer Vision and Pattern Recognition (CVPR)}, 2024.

\bibitem[Cioppa et~al.(2022)Cioppa, Giancola, Deliege, Kang, Zhou, Cheng, Ghanem, and Van~Droogenbroeck]{soccernet-tracking2022_ref}
Anthony Cioppa, Silvio Giancola, Adrien Deliege, Le Kang, Xin Zhou, Zhiyu Cheng, Bernard Ghanem, and Marc Van~Droogenbroeck.
\newblock Soccernet-tracking: Multiple object tracking dataset and benchmark in soccer videos.
\newblock In \emph{Proceedings of the IEEE/CVF Conference on Computer Vision and Pattern Recognition}, pages 3491--3502, 2022.

\bibitem[Cui et~al.(2023)Cui, Zeng, Zhao, Yang, Wu, and Wang]{sportsmot_ref}
Yutao Cui, Chenkai Zeng, Xiaoyu Zhao, Yichun Yang, Gangshan Wu, and Limin Wang.
\newblock Sportsmot: A large multi-object tracking dataset in multiple sports scenes.
\newblock \emph{Proceedings of the IEEE/CVF International Conference on Computer Vision (ICCV)}, 2023.

\bibitem[Dendorfer et~al.(2021)Dendorfer, Ossep, Milan, Cremers, Reid, Roth, and Leal-Taixe]{motchallenge_paper_ref}
Patrick Dendorfer, Aljossa Ossep, Anton Milan, Daniel Cremers, Ian Reid, Stefan Roth, and Laura Leal-Taixe.
\newblock Motchallenge: A benchmark for single-camera multiple target tracking.
\newblock \emph{International Journal of Computer Vision}, 129:\penalty0 1--37, 2021.

\bibitem[Du et~al.(2021)Du, Wan, Zhao, Zhang, Tong, and Dong]{nsa_kf_ref}
Yunhao Du, Junfeng Wan, Yanyun Zhao, Binyu Zhang, Zhihang Tong, and Junhao Dong.
\newblock Giaotracker: A comprehensive framework for mcmot with global information and optimizing strategies in visdrone 2021.
\newblock In \emph{Proceedings of the IEEE/CVF International Conference on Computer Vision (ICCV) Workshops}, pages 2809--2819, 2021.

\bibitem[Du et~al.(2023)Du, Zhao, Song, Zhao, Su, Gong, and Meng]{strongsort_ref}
Yunhao Du, Zhicheng Zhao, Yang Song, Yanyun Zhao, Fei Su, Tao Gong, and Hongying Meng.
\newblock Strongsort: Make deepsort great again.
\newblock \emph{IEEE Transactions on Multimedia}, 2023.

\bibitem[Gao and Wang(2023)]{memotr_ref}
Ruopeng Gao and Limin Wang.
\newblock {MeMOTR}: Long-term memory-augmented transformer for multi-object tracking.
\newblock In \emph{Proceedings of the IEEE/CVF International Conference on Computer Vision (ICCV)}, pages 9901--9910, 2023.

\bibitem[Gao et~al.(2024)Gao, Zhang, and Wang]{motip_ref}
Ruopeng Gao, Yijun Zhang, and Limin Wang.
\newblock Multiple object tracking as id prediction, 2024.

\bibitem[Ge et~al.(2021)Ge, Liu, Wang, Li, and Sun]{yolox_ref}
Zheng Ge, Songtao Liu, Feng Wang, Zeming Li, and Jian Sun.
\newblock Yolox: Exceeding yolo series in 2021.
\newblock \emph{arXiv preprint arXiv:2107.08430}, 2021.

\bibitem[He et~al.(2016)He, Zhang, Ren, and Sun]{resnet_ref}
Kaiming He, Xiangyu Zhang, Shaoqing Ren, and Jian Sun.
\newblock {Deep Residual Learning for Image Recognition}.
\newblock In \emph{Proceedings of 2016 IEEE Conference on Computer Vision and Pattern Recognition}, pages 770--778. IEEE, 2016.

\bibitem[Houlsby et~al.(2019)Houlsby, Giurgiu, Jastrzebski, Morrone, De~Laroussilhe, Gesmundo, Attariyan, and Gelly]{adapters_ref}
Neil Houlsby, Andrei Giurgiu, Stanislaw Jastrzebski, Bruna Morrone, Quentin De~Laroussilhe, Andrea Gesmundo, Mona Attariyan, and Sylvain Gelly.
\newblock Parameter-efficient transfer learning for {NLP}.
\newblock In \emph{Proceedings of the 36th International Conference on Machine Learning}, pages 2790--2799. PMLR, 2019.

\bibitem[Kalman(1960)]{kf_ref}
R.~E. Kalman.
\newblock {A New Approach to Linear Filtering and Prediction Problems}.
\newblock \emph{Journal of Basic Engineering}, 82\penalty0 (1):\penalty0 35--45, 1960.

\bibitem[Kirillov et~al.(2023)Kirillov, Mintun, Ravi, Mao, Rolland, Gustafson, Xiao, Whitehead, Berg, Lo, Doll{\'a}r, and Girshick]{sam_ref}
Alexander Kirillov, Eric Mintun, Nikhila Ravi, Hanzi Mao, Chloe Rolland, Laura Gustafson, Tete Xiao, Spencer Whitehead, Alexander~C. Berg, Wan-Yen Lo, Piotr Doll{\'a}r, and Ross Girshick.
\newblock Segment anything.
\newblock \emph{Proceedings of the IEEE/CVF International Conference on Computer Vision (ICCV)}, 2023.

\bibitem[Kuhn(1955)]{hungarianalg_ref}
H.~W. Kuhn.
\newblock The hungarian method for the assignment problem.
\newblock \emph{Naval Research Logistics Quarterly}, 2\penalty0 (1-2):\penalty0 83--97, 1955.

\bibitem[Li et~al.(2024)Li, Ke, Danelljan, Piccinelli, Segu, Van~Gool, and Yu]{masa_ref}
Siyuan Li, Lei Ke, Martin Danelljan, Luigi Piccinelli, Mattia Segu, Luc Van~Gool, and Fisher Yu.
\newblock Matching anything by segmenting anything.
\newblock \emph{CVPR}, 2024.

\bibitem[Lin et~al.(2014)Lin, Maire, Belongie, Bourdev, Girshick, Hays, Perona, Ramanan, Dollár, and Zitnick]{coco_dataset_ref}
Tsung-Yi Lin, Michael Maire, Serge~J. Belongie, Lubomir~D. Bourdev, Ross~B. Girshick, James Hays, Pietro Perona, Deva Ramanan, Piotr Dollár, and C.~Lawrence Zitnick.
\newblock Microsoft coco: Common objects in context.
\newblock \emph{CoRR}, abs/1405.0312, 2014.

\bibitem[Liu et~al.(2023)Liu, Zeng, Ren, Li, Zhang, Yang, Li, Yang, Su, Zhu, et~al.]{gr_dino_ref}
Shilong Liu, Zhaoyang Zeng, Tianhe Ren, Feng Li, Hao Zhang, Jie Yang, Chunyuan Li, Jianwei Yang, Hang Su, Jun Zhu, et~al.
\newblock Grounding dino: Marrying dino with grounded pre-training for open-set object detection.
\newblock \emph{arXiv preprint arXiv:2303.05499}, 2023.

\bibitem[Liu et~al.(2021)Liu, Lin, Cao, Hu, Wei, Zhang, Lin, and Guo]{swin_transformer_ref}
Ze Liu, Yutong Lin, Yue Cao, Han Hu, Yixuan Wei, Zheng Zhang, Stephen Lin, and Baining Guo.
\newblock Swin transformer: Hierarchical vision transformer using shifted windows.
\newblock In \emph{Proceedings of the IEEE/CVF International Conference on Computer Vision (ICCV)}, 2021.

\bibitem[Luiten et~al.(2020)Luiten, Osep, Dendorfer, Torr, Geiger, Leal-Taix{\'e}, and Leibe]{hota_ref}
Jonathon Luiten, Aljosa Osep, Patrick Dendorfer, Philip Torr, Andreas Geiger, Laura Leal-Taix{\'e}, and Bastian Leibe.
\newblock Hota: A higher order metric for evaluating multi-object tracking.
\newblock \emph{International Journal of Computer Vision}, pages 1--31, 2020.

\bibitem[Lv et~al.(2024)Lv, Huang, Zhang, Lin, Han, and Zeng]{diffmot_ref}
Weiyi Lv, Yuhang Huang, Ning Zhang, Ruei-Sung Lin, Mei Han, and Dan Zeng.
\newblock Diffmot: A real-time diffusion-based multiple object tracker with non-linear prediction.
\newblock In \emph{Proceedings of the IEEE/CVF Conference on Computer Vision and Pattern Recognition}, pages 19321--19330, 2024.

\bibitem[Maggiolino et~al.(2023)Maggiolino, Ahmad, Cao, and Kitani]{deepocsort_ref}
Gerard Maggiolino, Adnan Ahmad, Jinkun Cao, and Kris Kitani.
\newblock Deep oc-sort: Multi-pedestrian tracking by adaptive re-identification.
\newblock \emph{arXiv preprint arXiv:2302.11813}, 2023.

\bibitem[Milan et~al.(2016)Milan, Leal-Taix\'{e}, Reid, Roth, and Schindler]{mot17_ref}
A. Milan, L. Leal-Taix\'{e}, I. Reid, S. Roth, and K. Schindler.
\newblock {MOT}16: {A} benchmark for multi-object tracking.
\newblock \emph{arXiv:1603.00831 [cs]}, 2016.
\newblock arXiv: 1603.00831.

\bibitem[Qin et~al.(2024)Qin, Wang, Zhou, Fu, Hua, and Tang]{generaltrack_ref}
Zheng Qin, Le Wang, Sanping Zhou, Panpan Fu, Gang Hua, and Wei Tang.
\newblock Towards generalizable multi-object tracking.
\newblock In \emph{Proceedings of the IEEE/CVF Conference on Computer Vision and Pattern Recognition (CVPR)}, pages 18995--19004, 2024.

\bibitem[Ravi et~al.(2024)Ravi, Gabeur, Hu, Hu, Ryali, Ma, Khedr, R{\"a}dle, Rolland, Gustafson, Mintun, Pan, Alwala, Carion, Wu, Girshick, Doll{\'a}r, and Feichtenhofer]{sam2_ref}
Nikhila Ravi, Valentin Gabeur, Yuan-Ting Hu, Ronghang Hu, Chaitanya Ryali, Tengyu Ma, Haitham Khedr, Roman R{\"a}dle, Chloe Rolland, Laura Gustafson, Eric Mintun, Junting Pan, Kalyan~Vasudev Alwala, Nicolas Carion, Chao-Yuan Wu, Ross Girshick, Piotr Doll{\'a}r, and Christoph Feichtenhofer.
\newblock Sam 2: Segment anything in images and videos.
\newblock \emph{arXiv preprint arXiv:2408.00714}, 2024.

\bibitem[Ristani et~al.(2016)Ristani, Solera, Zou, Cucchiara, and Tomasi]{idf1_ref}
Ergys Ristani, Francesco Solera, Roger Zou, Rita Cucchiara, and Carlo Tomasi.
\newblock Performance measures and a data set for multi-target, multi-camera tracking.
\newblock In \emph{Computer Vision -- ECCV 2016 Workshops}, pages 17--35, Cham, 2016. Springer International Publishing.

\bibitem[Rublee et~al.(2011)Rublee, Rabaud, Konolige, and Bradski]{orb_ref}
Ethan Rublee, Vincent Rabaud, Kurt Konolige, and Gary Bradski.
\newblock Orb: an efficient alternative to sift or surf.
\newblock pages 2564--2571, 2011.

\bibitem[Sun et~al.(2021)Sun, Cao, Jiang, Yuan, Bai, Kitani, and Luo]{dancetrack_ref}
Peize Sun, Jinkun Cao, Yi Jiang, Zehuan Yuan, Song Bai, Kris Kitani, and Ping Luo.
\newblock Dancetrack: Multi-object tracking in uniform appearance and diverse motion.
\newblock \emph{Proceedings of the IEEE/CVF Conference on Computer Vision and Pattern Recognition (CVPR)}, 2021.

\bibitem[Yang et~al.(2022)Yang, Odashima, Masui, and Jiang]{cbiou_tech_report_ref}
Fan Yang, Shigeyuki Odashima, Shoichi Masui, and Shan Jiang.
\newblock The second-place solution for cvpr 2022 soccernet tracking challenge.
\newblock \emph{arXiv preprint arXiv:2211.13481}, 2022.

\bibitem[Yang et~al.(2023)Yang, Odashima, Masui, and Jiang]{cbiou_ref}
Fan Yang, Shigeyuki Odashima, Shoichi Masui, and Shan Jiang.
\newblock Hard to track objects with irregular motions and similar appearances? make it easier by buffering the matching space.
\newblock In \emph{Proceedings of the IEEE/CVF Winter Conference on Applications of Computer Vision (WACV)}, pages 4799--4808, 2023.

\bibitem[Yang et~al.(2024)Yang, Han, Yan, Zhang, Qi, Lu, and Wang]{hybridsort_ref}
Mingzhan Yang, Guangxin Han, Bin Yan, Wenhua Zhang, Jinqing Qi, Huchuan Lu, and Dong Wang.
\newblock Hybrid-sort: Weak cues matter for online multi-object tracking.
\newblock In \emph{Proceedings of the AAAI Conference on Artificial Intelligence}, pages 6504--6512, 2024.

\bibitem[Yu et~al.(2022)Yu, Li, Han, and Wang]{relationtrack_ref}
En Yu, Zhuoling Li, Shoudong Han, and Hongwei Wang.
\newblock Relationtrack: Relation-aware multiple object tracking with decoupled representation.
\newblock \emph{IEEE Transactions on Multimedia}, 25:\penalty0 2686–2697, 2022.

\bibitem[Zeng et~al.(2022)Zeng, Dong, Zhang, Wang, Zhang, and Wei]{motr_ref}
Fangao Zeng, Bin Dong, Yuang Zhang, Tiancai Wang, Xiangyu Zhang, and Yichen Wei.
\newblock Motr: End-to-end multiple-object tracking with transformer.
\newblock In \emph{European Conference on Computer Vision (ECCV)}, 2022.

\bibitem[Zhang et~al.(2021)Zhang, Wang, Wang, Zeng, and Liu]{fairmot_ref}
Yifu Zhang, Chunyu Wang, Xinggang Wang, Wenjun Zeng, and Wenyu Liu.
\newblock Fairmot: On the fairness of detection and re-identification in multiple object tracking.
\newblock \emph{International Journal of Computer Vision}, 129:\penalty0 3069--3087, 2021.

\bibitem[Zhang et~al.(2022)Zhang, Sun, Jiang, Yu, Weng, Yuan, Luo, Liu, and Wang]{bt_ref}
Yifu Zhang, Peize Sun, Yi Jiang, Dongdong Yu, Fucheng Weng, Zehuan Yuan, Ping Luo, Wenyu Liu, and Xinggang Wang.
\newblock Bytetrack: Multi-object tracking by associating every detection box.
\newblock 2022.

\bibitem[Zhang et~al.(2023)Zhang, Wang, and Zhang]{motrv2_ref}
Yuang Zhang, Tiancai Wang, and Xiangyu Zhang.
\newblock Motrv2: Bootstrapping end-to-end multi-object tracking by pretrained object detectors.
\newblock In \emph{Proceedings of the IEEE/CVF Conference on Computer Vision and Pattern Recognition (CVPR)}, pages 22056--22065, 2023.

\bibitem[Zhou et~al.(2020)Zhou, Koltun, and Kr{\"a}henb{\"u}hl]{centertrack_ref}
Xingyi Zhou, Vladlen Koltun, and Philipp Kr{\"a}henb{\"u}hl.
\newblock Tracking objects as points.
\newblock \emph{Proceedings of the European Conference on Computer Vision (ECCV)}, 2020.

\bibitem[Zhou et~al.(2022)Zhou, Girdhar, Joulin, Kr{\"a}henb{\"u}hl, and Misra]{detic_ref}
Xingyi Zhou, Rohit Girdhar, Armand Joulin, Philipp Kr{\"a}henb{\"u}hl, and Ishan Misra.
\newblock Detecting twenty-thousand classes using image-level supervision.
\newblock In \emph{Proceedings of the European Conference on Computer Vision (ECCV)}, 2022.

\end{thebibliography}
}

\newpage

\appendix
\begin{appendices}

This supplementary material contains the following appendices as referred in the main paper:

\begin{itemize}
  \item ~\ref{sec:mask_based_systems_more_exps} More experiments and details with mask-based tracking systems
  \item ~\ref{sec:sota_transformers_sushi} State-of-the-art comparison with transformer-based and other types of method
  % \item ~\ref{sec:cbiou_more_in_detail} Our C-BIoU implementation and its performance \textbf{\textcolor{red}{(TO BE DISCUSSED!)}}
  % \begin{itemize}
        % \item ~\ref{sec:cbiou_more_in_detail_first_subsec} Our C-BIoU implementation and its performance
        % \item ~\ref{sec:cbiou_more_in_detail_second_subsec} C-BIoU with temporally propagated mask as an association cue
    % \end{itemize}
  % \item ~\ref{sec:param_tuning} Parameter tuning and its impacts
  % \item ~\ref{sec:pub_dets} McByte and baseline on public detections \textbf{\textcolor{orange}{(To be removed)}}
  \item ~\ref{sec:app_f_more_visuals} Additional visual examples
  % \begin{itemize}
  %       \item ~\ref{sec:app_f_mask_management} Mask processing and handling \textbf{\textcolor{green}{(Updated)}}
  %       \item ~\ref{sec:app_f_mask_visib} Tracklet mask visibility at the scene
  %       \item ~\ref{sec:app_f_mask_conf} Mask confidence
  %       \item ~\ref{sec:app_f_mm1_mm2} Bounding box coverage (mm1) and mask fill ratio (mm2)
  %       \item ~\ref{sec:app_f_cmc} Camera motion compensation
  %       \item ~\ref{sec:app_f_more_visuals} Additional visual example \textbf{\textcolor{red}{[More than 1 extra figure added added - update the text!]}}
  %   \end{itemize}
  \item ~\ref{sec:app_speed_and_heaviness} The running speed and heaviness of mask
\end{itemize}

\begin{table*}
  \centering
  {\small{
  \scalebox{0.9}{
  \begin{tabular}{@{}lccc@{}}
    \toprule
    Details & HOTA & IDF1 & MOTA \\
    \midrule
    \multicolumn{4}{c}{DEVA} \\
    \midrule
    GDino "person", th. 0.35 $\ddagger$ & \textbf{31.8} & \textbf{31.3} & \textbf{-89.4} \\
    YOLOX ByteTrack, th. 0.6 $\ddagger$ & 24.7 & 20.4 & -239.7 \\
    YOLOX ByteTrack, th. 0.7 & 27.0 & 23.7 & -187.8 \\
    \midrule
    \multicolumn{4}{c}{Grounded SAM 2} \\
    \midrule
    GDino "person", th. 0.25, step 20 $\ddagger$ & 43.4 & 47.6 & 18.4 \\
    GDino "person", th. 0.25, step 100  & 44.0 & 49.0 & 15.5 \\
    YOLOX ByteTrack, th. 0.25, step 20 & 46.4 & 51.6 & 36.0 \\
    YOLOX ByteTrack, th. 0.6, step 20 $\ddagger$ & \textbf{47.5} & 54.1 & 43.0 \\
    YOLOX ByteTrack, th. 0.7, step 20 & 47.4 & 54.1 & 44.3 \\
    YOLOX ByteTrack, th. 0.25, step 100 & 46.8 & 54.2 & 30.2 \\
    YOLOX ByteTrack, th. 0.6, step 100 & 47.4 & \textbf{54.9} & 34.8 \\
    YOLOX ByteTrack, th. 0.7, step 100 & 47.4 & \textbf{54.9} & 35.9 \\
    YOLOX ByteTrack, th. 0.25, step 1 & 43.0 & 43.9 & 36.2 \\
    YOLOX ByteTrack, th. 0.6, step 1 & 44.4 & 46.5 & 44.9 \\
    YOLOX ByteTrack, th. 0.7, step 1 & 44.3 & 46.7 & \textbf{46.5} \\
    \midrule
    \multicolumn{4}{c}{MASA} \\
    \midrule
    GDino feat. Detic-SwinB "person", th 0.2 & 46.8 & 52.1 & 24.3 \\
    GDino feat. YOLOX COCO, th 0.3 & 45.4 & 53.1 & 36.9 \\
    GDino feat. YOLOX ByteTrack, th 0.3 & 61.8 & 70.8 & 71.3 \\
    GDino feat. YOLOX ByteTrack, th 0.6  & 63.4 & 73.3 & 73.8 \\
    GDino feat. YOLOX ByteTrack, th 0.7  & 62.5 & 71.9 & 72.9 \\
    R50 feat. YOLOX COCO, th 0.3 $\ddagger$ & 45.5 & 53.6 & 36.9 \\
    R50 feat. YOLOX ByteTrack, th 0.3 & 62.5 & 72.0 & 71.5 \\
    R50 feat. YOLOX ByteTrack, th 0.6 $\ddagger$ & \textbf{63.5} & \textbf{73.6} & \textbf{74.0} \\
    R50 feat. YOLOX ByteTrack, th 0.7 & 62.6 & 72.3 & 73.0 \\
    \midrule
    \multicolumn{4}{c}{McByte} \\
    \midrule
    McByte (ours)           & \textbf{69.9} & \textbf{82.8} & \textbf{78.5} \\
    \bottomrule
  \end{tabular}
  }
  }}
    \caption{Extended comparison with the other tracking methods using segmentation mask: DEVA~\cite{deva_ref}, Grounded SAM 2~\cite{sam_ref,gr_dino_ref} and MASA~\cite{masa_ref} on MOT17 validation set~\cite{mot17_ref}, while changing their parameters. $\ddagger$ denotes the variants reported in the main paper and in ~\cref{tab:deva_grsam2_masa_dt_val}.}
  \label{tab:deva_grsam2_masa_mot17val_extended}
\end{table*}

\begin{table}
  \centering
  {\small{
  \scalebox{0.9}{
  \begin{tabular}{@{}lccc@{}}
    \toprule
    Method & HOTA & IDF1 & MOTA \\
    \midrule
    DEVA, original settings        & 21.9 & 15.8 & -347.1 \\
    DEVA, with YOLOX               & 20.1 & 13.3 & -423.9 \\
    \midrule
    Grounded SAM 2, original settings & 51.3 & 48.0 & 73.5 \\
    Grounded SAM 2, with YOLOX        & 52.9 & 49.6 & 81.6 \\
    \midrule
    MASA, original settings           & 38.2 & 34.9 & 71.9 \\
    MASA, with YOLOX                  & 46.0 & 41.1 & 85.6 \\
    \midrule
    McByte (ours)                     & \textbf{62.3} & \textbf{64.0} & \textbf{89.8} \\
    \bottomrule
  \end{tabular}
  }
  }}
  \caption{Comparison with the other tracking methods using segmentation mask: DEVA~\cite{deva_ref}, Grounded SAM 2~\cite{sam_ref,gr_dino_ref} and MASA~\cite{masa_ref} on DanceTrack validation set~\cite{dancetrack_ref}. The reported variants correspond to the variants with $\ddagger$ symbol in ~\cref{tab:deva_grsam2_masa_mot17val_extended}}
  % \vspace*{-0.4cm}
  \label{tab:deva_grsam2_masa_dt_val}
\end{table}

\section{More experiments and details with mask-based tracking systems}
\label{sec:mask_based_systems_more_exps}

%%% More experiments on DEVA, Gr SAM 2 and MASA %%%

%%% / %%%

% We evaluate DEVA~\cite{deva_ref}, Grounded SAM 2~\cite{sam2_ref, gr_dino_ref} and MASA~\cite{masa_ref} on MOT datasets. For every outputted bounding box on each frame, we save its data to a text file following the MOT format convention~\cite{motchallenge_paper_ref}. While Grounded SAM 2 and MASA provide IDs of the tracked objects, in case of DEVA, we use immutable and unique IDs of the propagated masks associated with each bounding box. 
%
We evaluate DEVA~\cite{deva_ref}, Grounded SAM 2~\cite{sam2_ref, gr_dino_ref}, and MASA~\cite{masa_ref} on MOT datasets, saving each bounding box output per frame in MOT format~\cite{motchallenge_paper_ref}. 
% Grounded SAM 2 and MASA provide object IDs, while for DEVA, we use unique, immutable IDs from the propagated masks associated with each bounding box.
% The limitations of these methods when applied to MOT are described in Sec. 2.3 and 4.6 of the main paper.

% We provide more experiments to demonstrate careful exploration of the mask-based tracking systems and their performance differences with respect to our McByte. We provide more variants on MOT17~\cite{mot17_ref} validation set and an analogous table as in the main paper, yet on DanceTrack~\cite{dancetrack_ref} validation set.
%
We conduct additional experiments to thoroughly explore the performance differences between the mask-based tracking systems and our McByte. These include several variants on the MOT17~\cite{mot17_ref} validation set, as well as experiments on the DanceTrack~\cite{dancetrack_ref} validation set, analogous to the ones presented in the main paper.

% \cref{tab:deva_grsam2_masa_mot17val_extended} lists different variants run on MOT17 validation set, where we change the detectors and used parameters. $\ddagger$ denotes the variants that are included in the main paper.
%
\cref{tab:deva_grsam2_masa_mot17val_extended} presents various experimental variants on the MOT17 validation set, where different detectors and parameters are used. The variants marked with $\ddagger$ correspond to those discussed in the main paper on SportsMOT~\cite{sportsmot_ref}.

% For DEVA, we first run it with the default settings, i.e. with the Grounding Dino~\cite{gr_dino_ref} detector with prompt "person". The confidence threshold for not suppressing the bounding boxes is 0.35. Next, we insert the YOLOX~\cite{yolox_ref} detector with the weights trained on the MOT17 dataset, provided by our baseline~\cite{bt_ref}. We run variants with two threshold: 0.6 and 0.7, which correspond to the values of high confidence detection threshold and new tracklet initialization in our method respectively.
%
For DEVA, we first run the default settings using the Grounding Dino~\cite{gr_dino_ref} detector with the "person" prompt and a confidence threshold of 0.35 to accept bounding boxes. Then, we replace it with the YOLOX~\cite{yolox_ref} detector, trained on the MOT17 dataset from our baseline~\cite{bt_ref}. We test two threshold values, 0.6 and 0.7. In our baseline, initialization of the new tracklets happens for the values 0.1 higher than the high confidence detection threshold. As we consider the default value of 0.6 for the latter (\cref{sec:exps_impl_details} in the main paper), we also experiment with the value of 0.7 with DEVA and other mask based systems.
% which correspond to the high-confidence detection threshold \textbf{\textcolor{red}{[REMOVE IT, for consistency with the main paper] and the new tracklet initialization}} in our method, respectively (Sec. 4.1 in the main paper).

% For Grounded SAM 2~\cite{sam2_ref,gr_dino_ref}, we consider the version of "Video Object Tracking with Continues ID" as specified on its github page\footnote{\url{https://github.com/IDEA-Research/Grounded-SAM-2}}. We first run the variant with original settings: using Grounding Dino~\cite{gr_dino_ref} detector with prompt "person", original confidence detection threshold 0.25 and the step value of 20. Step determines how often the detections are processed (e.g. every 20th frame) and objects are established to create the mask tracklets. It is considered as the segment length (recall tracking objects in segments mentioned in the main paper). Next, we run the analogous variant with step value of 100. 
%
For Grounded SAM 2~\cite{sam2_ref,gr_dino_ref}, we use the "Video Object Tracking with Continuous ID" version as specified on its GitHub page\footnote{\url{https://github.com/IDEA-Research/Grounded-SAM-2}}. Initially, we run it with the original settings, using the Grounding Dino~\cite{gr_dino_ref} detector with the "person" prompt, a confidence detection threshold of 0.25, and a step value of 20. The step value defines how often detections are processed (e.g., every 20th frame) to create mask tracklets, functioning as the segment length (we refer to tracking objects in segments mentioned in the main paper, \cref{sec:exps_other_mask_methods}). We then test an analogous variant with a step value of 100.

% Further, we insert the YOLOX detectors with the weights from our baseline~\cite{bt_ref} and run variants with the step values of 20, 100 and 1 with different bounding box allowance thresholds 0.25 as well as 0.6 and 0.7 (as mentioned for DEVA). We also attempt to run the variant with the segment lenght of the whole video sequence, however it cannot be finished due to the extensive GPU memory requirement. Besides, it would only track the objects appearing during the first frame.
%
Next, we integrate YOLOX detector with weights from our baseline~\cite{bt_ref} and run variants with step values of 20, 100, and 1 (thus processing detections every frame), using different bounding box allowance thresholds of 0.25, 0.6, and 0.7 (analogous to the DEVA experiments). We also attempt to run a variant with the segment length set to the entire video sequence, but it fails due to excessive GPU memory requirements. Additionally, this setup would only track objects visible in the first frame.

% MASA~\cite{masa_ref} provides a few models for inference. We run variants with two different feature backbones: GroundingDINO~\cite{gr_dino_ref} (GDino) and ResNet-50~\cite{resnet_ref} (R50). We run the GroundingDINO variant with Detic-SwinB detector~\cite{detic_ref,swin_transformer_ref} with prompt "person" with the original detection confidence threshold of 0.2. We run an analogous variant, but with YOLOX detector trained on COCO~\cite{coco_dataset_ref} dataset as provided by the authors, with original threshold of 0.3.
%
MASA~\cite{masa_ref} offers several models for inference. We test variants using two different feature backbones: GroundingDINO~\cite{gr_dino_ref} (GDino) and ResNet-50~\cite{resnet_ref} (R50). For the GroundingDINO variant, we use the Detic-SwinB detector~\cite{detic_ref,swin_transformer_ref} with the "person" prompt, applying the original detection confidence threshold of 0.2. We also run a similar variant with the YOLOX detector trained on the COCO~\cite{coco_dataset_ref} dataset, as provided by the authors, using a confidence threshold of 0.3, default for this variant.

% Further, we insert YOLOX detector with the weights from our baseline~\cite{bt_ref} and run variants with different detection confidence thresholds: 0.3, 0.6 and 0.7 (as mentioned for DEVA and Grounded SAM 2). We also run the ResNet-50 feature variants with YOLOX COCO version (threshold 0.3) and weights as pre-trained by our baseline (thresholds 0.3, 0.6, 0.7).
%
Further, we incorporate the YOLOX detector with weights from our baseline~\cite{bt_ref} and test variants with detection confidence thresholds of 0.3, 0.6, and 0.7, analogously to DEVA and Grounded SAM 2. Additionally, we run the ResNet-50 feature variants with the YOLOX COCO model (threshold 0.3) and the baseline-pre-trained weights (thresholds 0.3, 0.6, 0.7).

% As it is demonstrated in \cref{tab:deva_grsam2_masa_mot17val_extended}, McByte performs better than the referenced mask-based systems, which makes it more suitable for MOT.
%
As shown in \cref{tab:deva_grsam2_masa_mot17val_extended}, McByte outperforms the referenced mask-based systems, making it more suitable for MOT.

% \cref{tab:deva_grsam2_masa_dt_val} lists the performance of DEVA, Grounded SAM 2 and MASA on DanceTrack~\cite{dancetrack_ref} validation set. The listed variants correspond to the ones with $\ddagger$ symbol in ~\cref{tab:deva_grsam2_masa_mot17val_extended} and thus to the variants reported in the main paper.
%
\cref{tab:deva_grsam2_masa_dt_val} presents the performance of DEVA, Grounded SAM 2, and MASA on the DanceTrack~\cite{dancetrack_ref} validation set. The listed variants correspond to those marked with $\ddagger$ in ~\cref{tab:deva_grsam2_masa_mot17val_extended} and are the ones reported in the main paper on SportsMOT.

% Also on DanceTrack, McByte manifests considerably higher performance demonstrating effectiveness and suitability for MOT.
% 
On DanceTrack, McByte also demonstrates significantly higher performance, reinforcing its effectiveness and suitability for MOT.

\section{State-of-the-art comparison with transformer-based and other types of method}
\label{sec:sota_transformers_sushi}

\begin{table}
  \centering
  {\small{
  \scalebox{0.9}{
  \begin{tabular}{@{}lccc@{}}
    \toprule
    Method & HOTA & IDF1 & MOTA \\
    \midrule
    \multicolumn{4}{c}{Transformer-based} \\
    \midrule
    MeMOTR~\cite{memotr_ref}            & 70.0 & 71.4 & 91.5 \\
    MOTIP~\cite{motip_ref}              & 71.9 & 75.0 & 92.9 \\
    \midrule
    \multicolumn{4}{c}{Joint detection and tracking} \\
    \midrule
    FairMOT~\cite{fairmot_ref}          & 49.3 & 53.5 & 86.4 \\
    CenterTrack~\cite{centertrack_ref}          & 62.7 & 60.0 & 90.8 \\
    \midrule
    \multicolumn{4}{c}{Tracking-by-detection} \\
    \midrule
    ByteTrack~\cite{bt_ref} & 64.1 & 71.4 & 95.9 \\
    MixSort-Byte~\cite{sportsmot_ref} & 65.7 & 74.1 & 96.2 \\
    OC-SORT~\cite{ocsort_ref} & 73.7 & 74.0 & 96.5 \\
    MixSort-OC~\cite{sportsmot_ref} & 74.1 & 74.4 & 96.5 \\
    GeneralTrack~\cite{generaltrack_ref} & 74.1 & 76.4 & 69.8 \\
    DiffMOT~\cite{diffmot_ref} & 76.2 & 76.1 & 97.1 \\
    McByte (ours)                 & 76.9 & 77.5 & 97.2 \\
    %%%
    % \midrule
    % \multicolumn{4}{c}{\textbf{\textcolor{red}{--- POSSIBLY REMOVE THIS PAPER ---}}} \\
    % \multicolumn{4}{c}{\textcolor{magenta}{WACV24 workshop -} \textbf{\textcolor{cyan}{CHECK THE COMMENT}} \textcolor{magenta}{in main paper}} \\
    % \midrule
    % \textcolor{magenta}{\st{Deep-EIoU (train)}} & \textcolor{magenta}{\st{74.1}} & \textcolor{magenta}{\st{75.0}} & \textcolor{magenta}{\st{95.1}} \\
    % \textbf{\textcolor{magenta}{Deep-EIoU (train+val)}} & \textbf{\textcolor{magenta}{77.2}} & \textbf{\textcolor{magenta}{79.8}} & \textcolor{magenta}{96.3} \\
    %%%             
    \bottomrule
  \end{tabular}
  }
    }}
  \caption{Extended state-of-the-art method comparison on SportsMOT~\cite{sportsmot_ref} test set.}
  % \vspace*{-0.3cm}
  \label{tab:sota_sportsmot_test_extended}
\end{table}

\begin{table}
  \centering
  {\small{
  \scalebox{0.9}{
  \begin{tabular}{@{}lccc@{}}
    \toprule
    Method & HOTA & IDF1 & MOTA \\
    \midrule
    \multicolumn{4}{c}{Transformer-based} \\
    \midrule
    MOTR~\cite{motr_ref}                & 57.8 & 68.6 & 73.4 \\
    MeMOTR~\cite{memotr_ref}            & 58.8 & 71.5 & 72.8 \\
    MOTRv2~\cite{motrv2_ref}            & 62.0 & 75.0 & 78.6 \\
    MOTIP~\cite{motip_ref}              & 59.2 & 71.2 & 75.5 \\
    \midrule
    \multicolumn{4}{c}{Global optimization} \\
    \midrule
    SUSHI~\cite{sushi_ref}              & 66.5 & 83.1 & 81.1 \\
    \midrule
    \multicolumn{4}{c}{Joint detection and tracking} \\
    \midrule
    FairMOT~\cite{fairmot_ref}          & 59.3 & 72.3 & 73.7 \\
    RelationTrack~\cite{relationtrack_ref}          & 61.0 & 75.8 & 75.6 \\
    CenterTrack~\cite{centertrack_ref}          & 52.2 & 64.7 & 67.8 \\
    \midrule
    \multicolumn{4}{c}{Tracking-by-detection } \\
    \multicolumn{4}{c}{with parameter tuning per sequence} \\
    \midrule
    \textcolor{gray}{ByteTrack~\cite{bt_ref}}             & \textcolor{gray}{63.1} & \textcolor{gray}{77.3} & \textcolor{gray}{80.3} \\
    \textcolor{gray}{MixSort-Byte~\cite{sportsmot_ref}}             & \textcolor{gray}{64.0} & \textcolor{gray}{78.7} & \textcolor{gray}{79.3} \\
    % BoT-SORT~\cite{botsort_ref}           & 65.0 & 80.2 & 80.5 \\
    \textcolor{gray}{StrongSORT++~\cite{strongsort_ref}}  & \textcolor{gray}{64.4} & \textcolor{gray}{79.5} & \textcolor{gray}{79.6} \\
    % \textcolor{violet}{C-BIoU~\cite{cbiou_ref}}          & \textcolor{violet}{64.1} & \textcolor{violet}{79.7} & \textcolor{violet}{81.1} \\
    % C-BIoU~\cite{cbiou_ref} *           & TO & BE & RUN \\
    % ImprAsso~\cite{imprasso_ref}        & \textbf{66.4} & \textbf{82.1} & \textbf{82.2} \\
    \textcolor{gray}{OC-SORT~\cite{ocsort_ref}}           & \textcolor{gray}{63.2} & \textcolor{gray}{77.5} & \textcolor{gray}{78.0} \\
    \textcolor{gray}{MixSort-OC~\cite{sportsmot_ref}}           & \textcolor{gray}{63.4} & \textcolor{gray}{77.8} & \textcolor{gray}{78.9} \\
    \textcolor{gray}{Deep OC-SORT~\cite{deepocsort_ref}}  & \textcolor{gray}{64.9} & \textcolor{gray}{80.6} & \textcolor{gray}{79.4} \\
    \textcolor{gray}{Hybrid-SORT~\cite{hybridsort_ref}}   & \textcolor{gray}{64.0} & \textcolor{gray}{78.7} & \textcolor{gray}{79.9} \\
    \midrule
    \multicolumn{4}{c}{Tracking-by-detection } \\
    \multicolumn{4}{c}{without parameter tuning per sequence} \\
    \midrule
    ByteTrack~\cite{sushi_ref} & 62.8 & 77.1 & 78.9 \\
    % \textcolor{blue}{C-BIoU~\cite{cbiou_ref} as reported (no code)}             & \textcolor{blue}{64.1} & \textcolor{blue}{\textbf{\underline{79.7}}} & \textcolor{blue}{\textbf{\underline{81.1}}} \\
    % C-BIoU~\cite{cbiou_ref} * our impl.          & 62.4 & 77.1 & 79.5 \\
    % \midrule
    % \multicolumn{4}{c}{Without parameter tuning per sequence} \\
    % \midrule
    % \textcolor{red}{[REMOVE IT]} \textcolor{magenta}{ByteTrack (with interp.)} & 63.2 & 77.5 & 79.7 \\
    % Ours (McByte)                    & 63.1 & 78.0 & 79.7 \\
    GeneralTrack ~\cite{generaltrack_ref} & 64.0 & 78.3 & 80.6 \\
    DiffMOT ~\cite{diffmot_ref} & 64.2 & 79.3 & 79.8 \\
    McByte (ours)                      & 64.2 & 79.4 & 80.2 \\
    % & \textbf{64.2} & \textbf{79.4} & \textbf{80.2} \\            
    \bottomrule
  \end{tabular}
  }
    }}
  \caption{Extended state-of-the-art method comparison on MOT17~\cite{mot17_ref} test set.}
  \vspace*{-0.3cm}
  \label{tab:sota_mot17_test_extended}
\end{table}
\begin{table}
  \centering
  {\small{
  \scalebox{0.9}{
  \begin{tabular}{@{}lccc@{}}
    \toprule
    Method & HOTA & IDF1 & MOTA \\
    \midrule
    \multicolumn{4}{c}{Transformer-based} \\
    \midrule
    MOTR~\cite{motr_ref}                & 54.2 & 51.5 & 79.7 \\
    MeMOTR~\cite{memotr_ref}            & 63.4 & 65.5 & 85.4 \\
    MOTRv2~\cite{motrv2_ref}            & 73.4 & 76.0 & 92.1 \\
    MOTIP~\cite{motip_ref}              & 67.5 & 72.2 & 90.3 \\
    \midrule
    \multicolumn{4}{c}{Global optimization} \\
    \midrule
    SUSHI~\cite{sushi_ref}              & 63.3 & 63.4 & 88.7 \\
    \midrule
    \multicolumn{4}{c}{Joint detection and tracking} \\
    \midrule
    FairMOT~\cite{fairmot_ref}          & 39.7 & 40.8 & 82.2 \\
    CenterTrack~\cite{centertrack_ref}          & 41.8 & 35.7 & 86.8 \\
    \midrule
    \multicolumn{4}{c}{Tracking-by-detection} \\
    \midrule
    ByteTrack~\cite{bt_ref}             & 47.7 & 53.9 & 89.6 \\
    MixSort-Byte~\cite{sportsmot_ref} & 46.7 & 53.0 & 85.5 \\
    OC-SORT~\cite{ocsort_ref}           & 55.1 & 54.9 & 92.2 \\
    % Deep OC-SORT~\cite{deepocsort_ref}  & 61.3 & 61.5 & 92.3 \\
    % \textcolor{blue}{C-BIoU~\cite{cbiou_ref} as reported (no code)}             & \textcolor{blue}{60.6} & \textcolor{blue}{61.6} & \textcolor{blue}{91.6} \\
    % C-BIoU~\cite{cbiou_ref} * our impl.          & 45.8 & 52.0 & 88.4 \\
    StrongSORT++~\cite{strongsort_ref}  & 55.6 & 55.2 & 91.1 \\
    Hybrid-SORT~\cite{hybridsort_ref}   & 65.7 & 67.4 & 91.8 \\
    GeneralTrack\cite{generaltrack_ref} & 59.2 & 59.7 & 91.8 \\
    DiffMOT~\cite{diffmot_ref} & 63.4 & 64.0 & 92.7 \\
    % Ours (McByte)            & 66.5 & \textbf{68.2} & 92.7 \\
    McByte (ours)                       & 67.1 & 68.1 & 92.9 \\             
    \bottomrule
  \end{tabular}
  }
    }}
  \caption{Extended state-of-the-art method comparison on DanceTrack~\cite{dancetrack_ref} test set.}
  \label{tab:sota_dancetrack_test_extended}
  \vspace*{-0.3cm}
\end{table}

%%% / %%%

% There are MOT methods outside the tracking-by-detection domain that perform better than ours on some benchmarks, 
There exist MOT methods outside the tracking-by-detection domain manifesting performance differences,
but usually these methods are not directly comparable, because they
require a lot of training data and might use other detections. Further,
they make certain hypotheses, e.g. global optimization on the whole video. At the same time, these methods might perform visibly worse on some benchmarks as we discuss below. On the contrary, we stress that McByte performs well on all the discussed benchmarks (\cref{sec:exps_abl_stud,sec:exps_sota_tr_by_det} of the main paper). McByte is a tracking-by-detection approach, which is the main focus of our work. For an additional reference, though, we also list performance of the transformer-based, global optimization, and joint detection and tracking methods. 

\cref{tab:sota_sportsmot_test_extended,tab:sota_dancetrack_test_extended,tab:sota_mot17_test_extended} show extended comparison including other types of tracking methods based on the result availability. All the tracking-by-detection methods use the same object detector models per dataset.

\cref{tab:sota_sportsmot_test_extended} presents extended state-of-the-art comparison on SportsMOT~\cite{sportsmot_ref} test set. In this dataset, the number of subjects can vary as due to abrupt camera motion, subjects can continuously enter and leave the scene. Further, due to the team sport nature, there are many occlusions and blur among the tracked objects. Transformer-based methods cannot handle all the mentioned challenges and perform lower than most of the tracking-by-detection approaches, including ours. Joint detection and tracking methods generalize poorly to this dataset and fall behind the other two types of tracking methods. Our method can handle the challenges present in the sport settings and outperforms all the other methods. 

\cref{tab:sota_mot17_test_extended} shows extended state-of-the art comparison on MOT17~\cite{mot17_ref} test set. Note that analogously to the main paper, we also put the result of ByteTrack~\cite{bt_ref} not being tuned per sequence as reported in \cite{sushi_ref} (\textit{"ByteTrack~\cite{sushi_ref}"}). Transformer-based methods perform visibly lower than the tracking-by-detection methods (including ours) as they struggle with the subjects frequently entering and leaving the scene. In contrast, SUSHI~\cite{sushi_ref}, which is a powerful global optimization approach, reaches highly satisfying performance. However, it accesses all the video frames at the same time while processing detections and associating the tracklets, which makes it impossible to run in online settings. Current state-of-the-art joint detection and tracking methods generally perform lower than the tracking-by-detection methods. In that paradigm, the detection and association step is performed jointly. In our method, we perform these two steps separately and focus on the association part.

\cref{tab:sota_dancetrack_test_extended} presents extended state-of-the-art comparison on DanceTrack~\cite{dancetrack_ref} test set. As in this dataset the subjects remain mostly at the scene, the transformer-based methods performance is more satisfying. The performance of transformer-based methods can be both higher~\cite{motrv2_ref,motip_ref} or lower~\cite{motr_ref,memotr_ref} compared to the the tracking-by-detection methods. For similar reasons, the global optimization method, SUSHI~\cite{sushi_ref} can also perform higher than the other tracking-by-detection methods on this dataset, or lower, e.g. when compared to our method. On this dataset, joint detection and tracking methods also manifest lower performance than the tracking-by-detection methods.

\section{Additional visual examples}
\label{sec:app_f_more_visuals}

\begin{figure*}
\centering
\begin{tabular}{ccc}
\includegraphics[height=3.0cm]{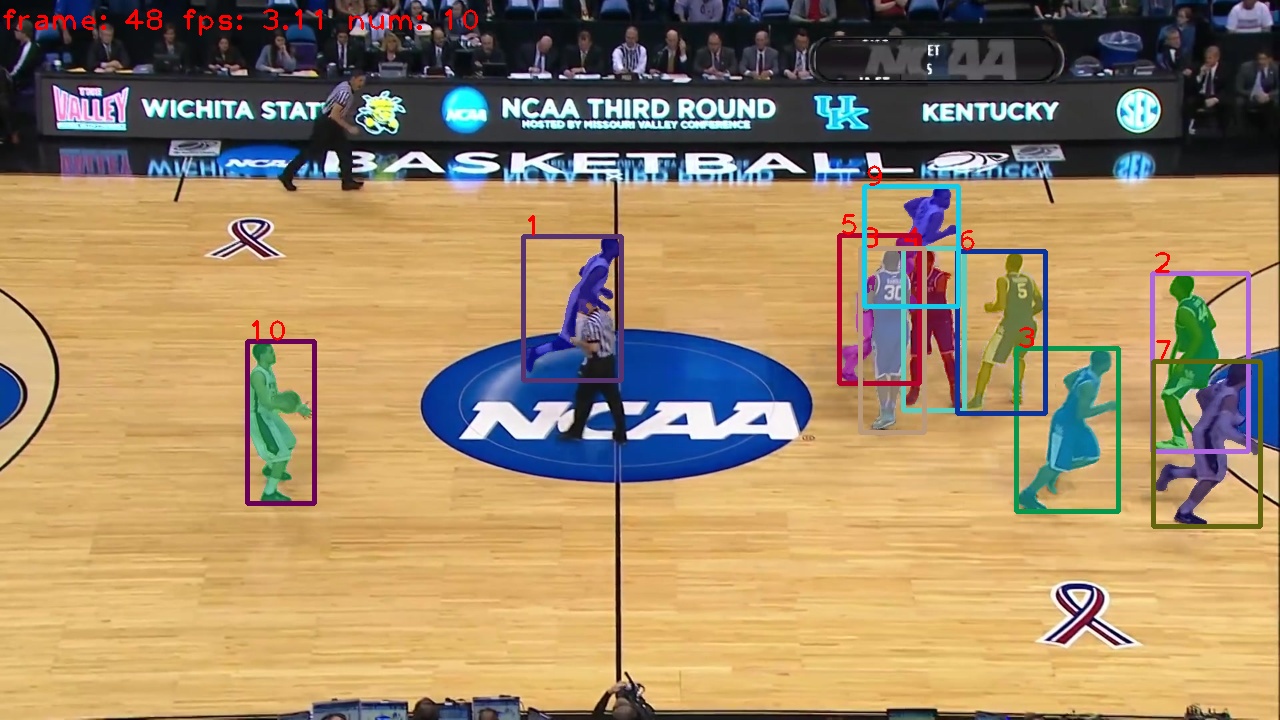}
\includegraphics[height=3.0cm]{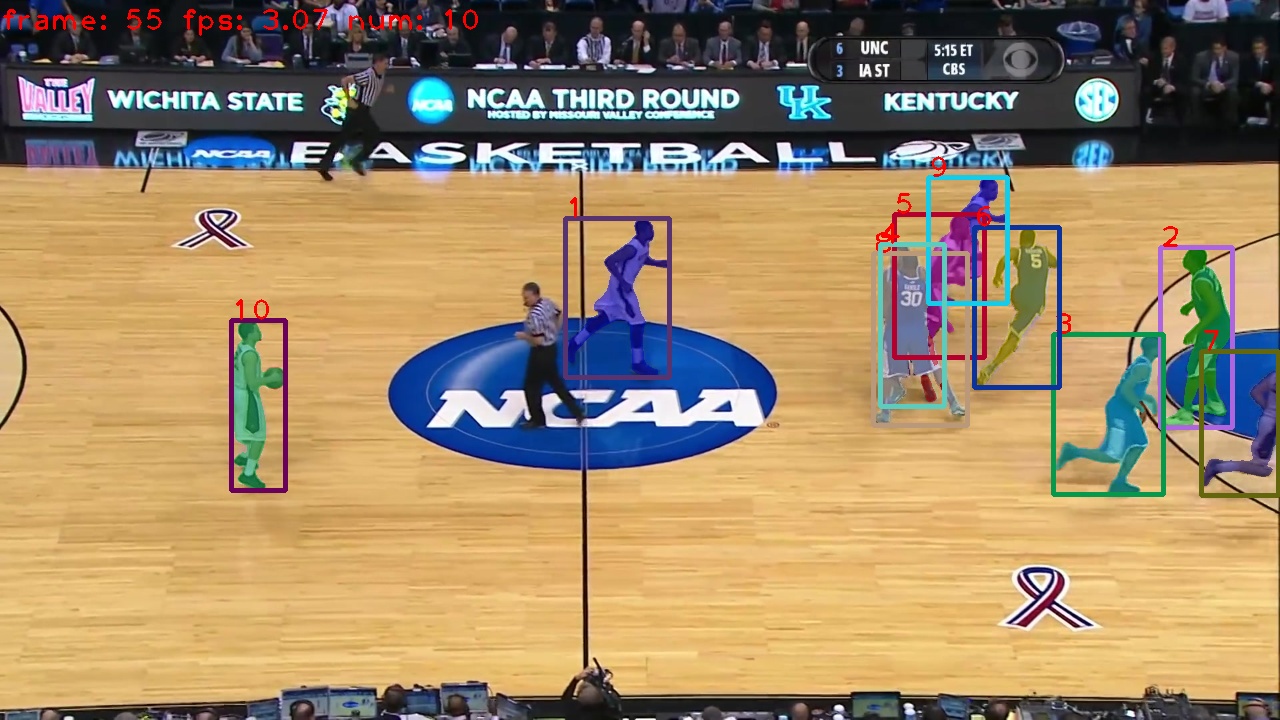}
\includegraphics[height=3.0cm]{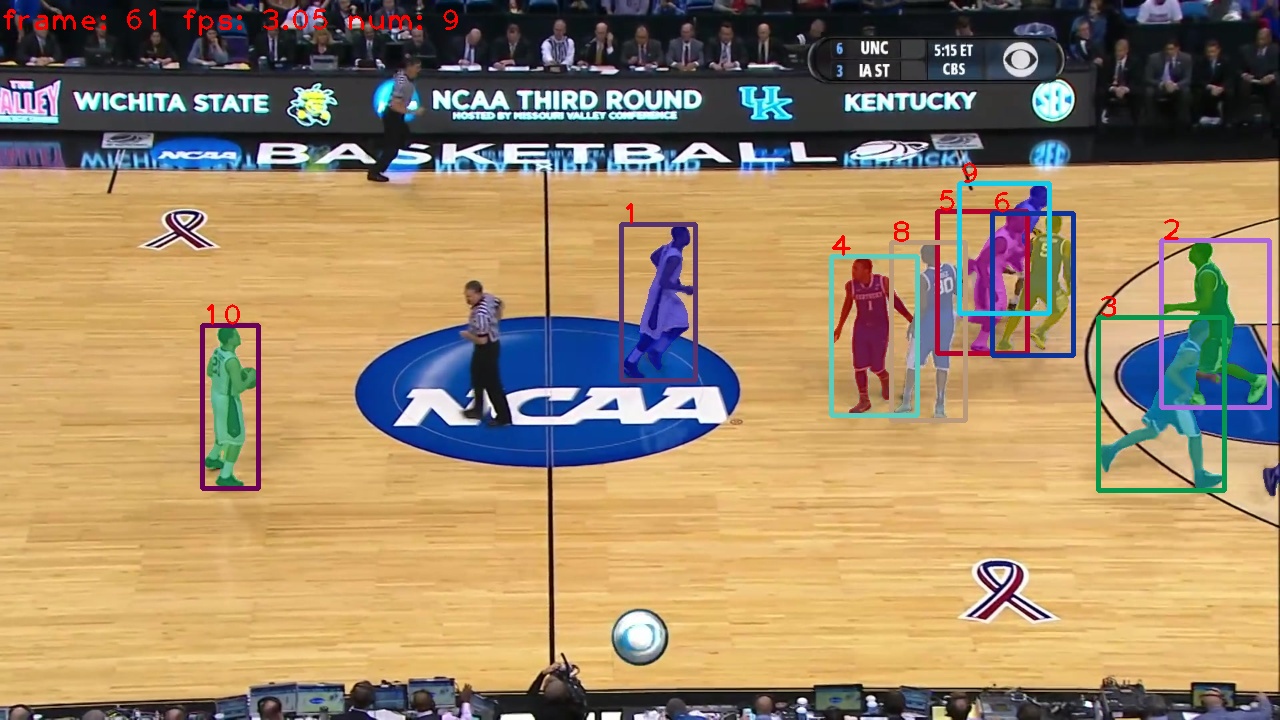}
\end{tabular}
\caption{
Full output frames corresponding to Fig. 1 from the main paper. Input image data from~\cite{sportsmot_ref}.
}
% \vspace*{-0.3cm}
\label{fig:full_basket_comp}
\end{figure*}

\begin{figure*}
\centering
\begin{tabular}{ccc}
Input frames  \\
\includegraphics[height=3.0cm]{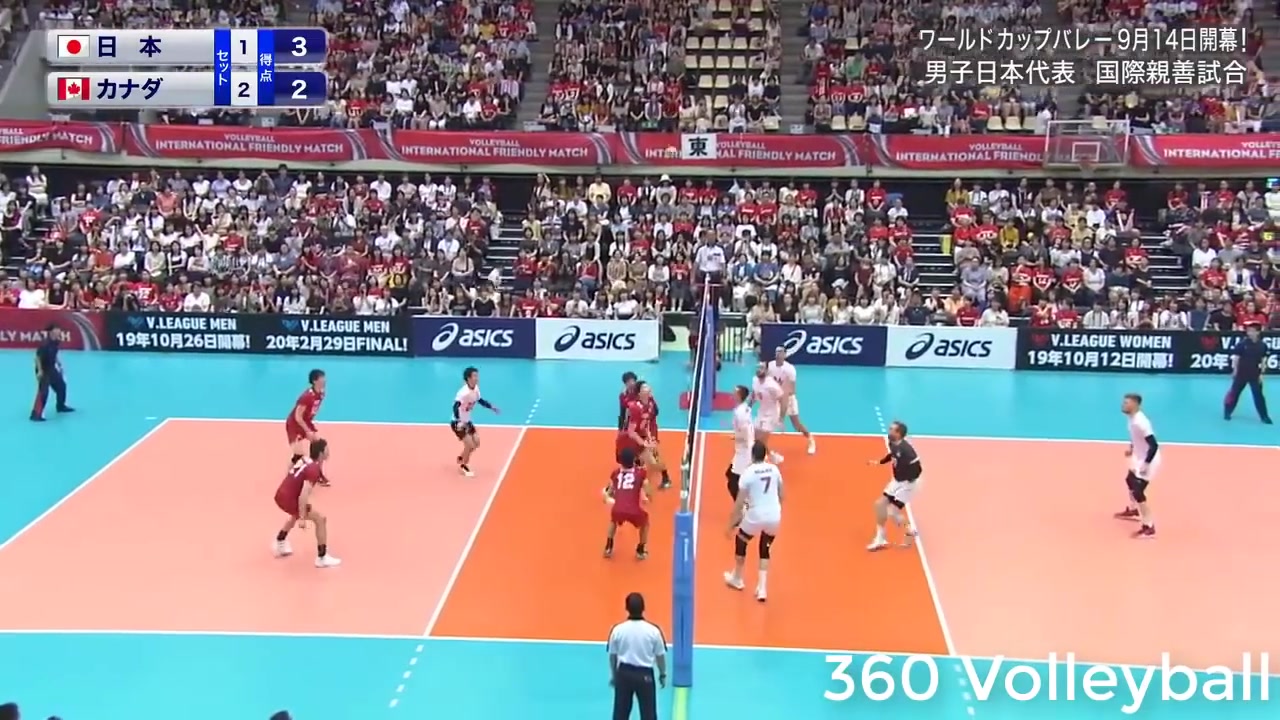}
\includegraphics[height=3.0cm]{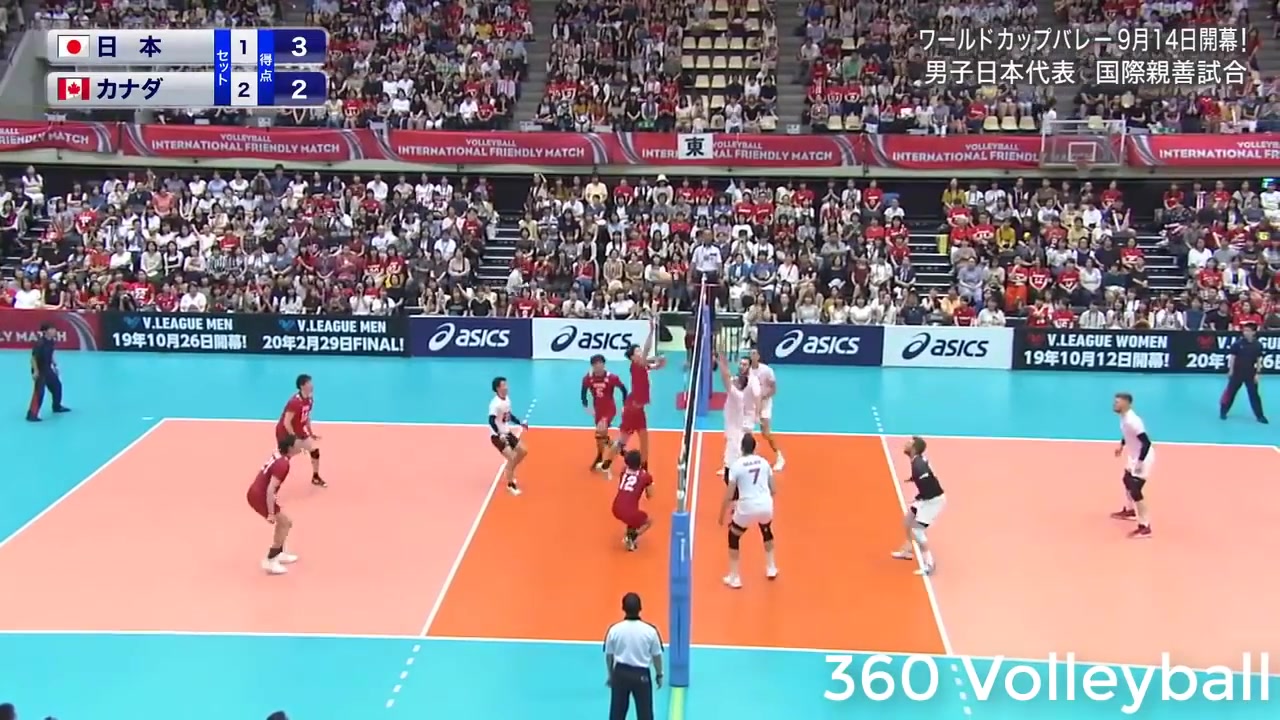}
\includegraphics[height=3.0cm]{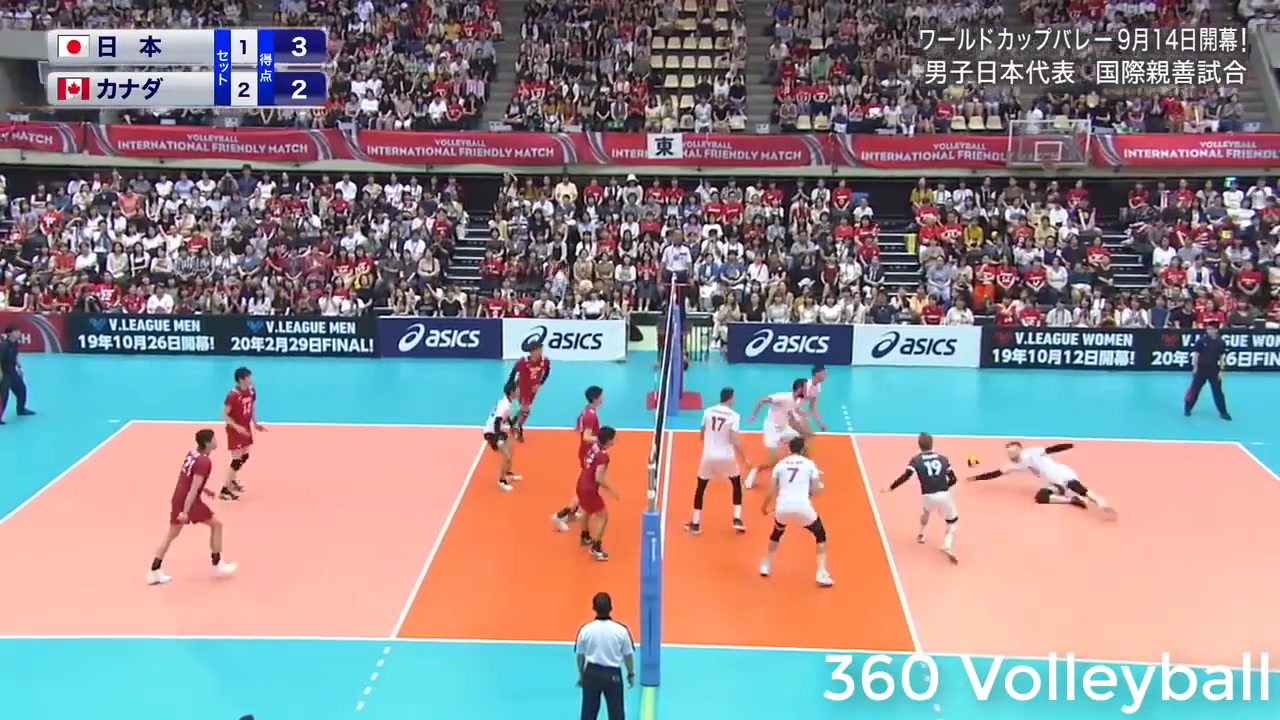}
\\
McByte (ours)  \\
\includegraphics[height=3.0cm]{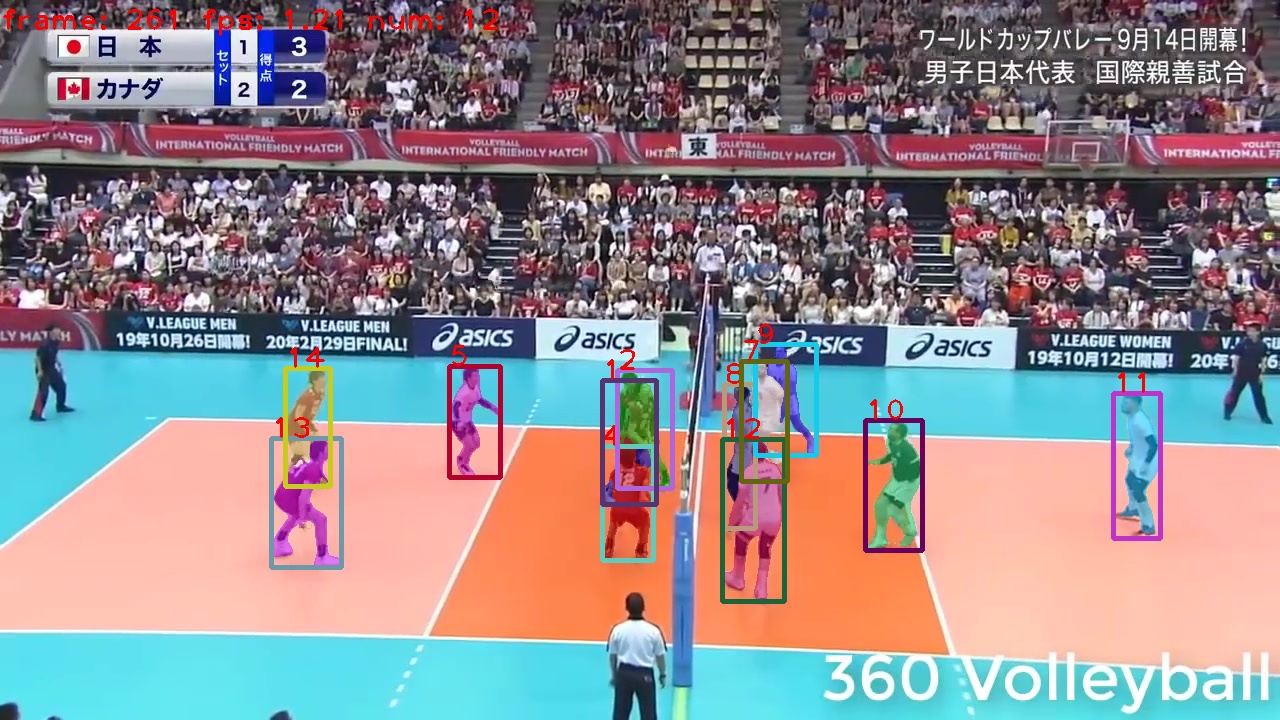}
\includegraphics[height=3.0cm]{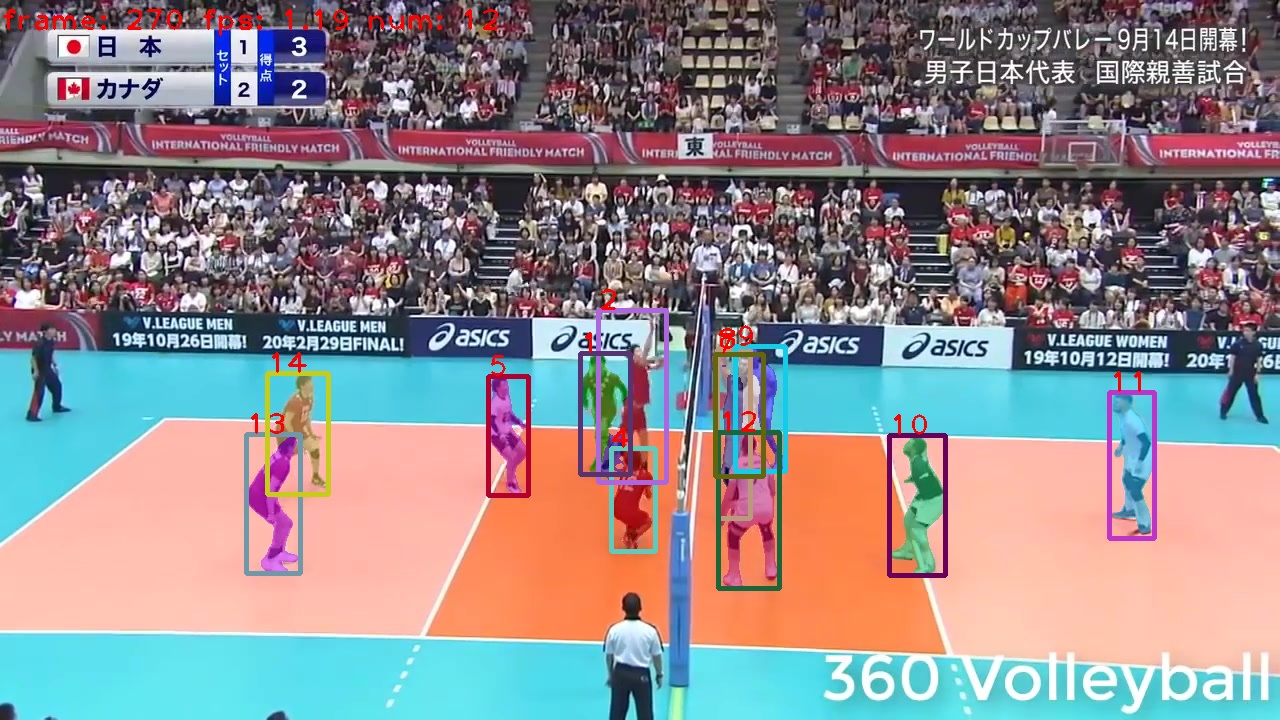}
\includegraphics[height=3.0cm]{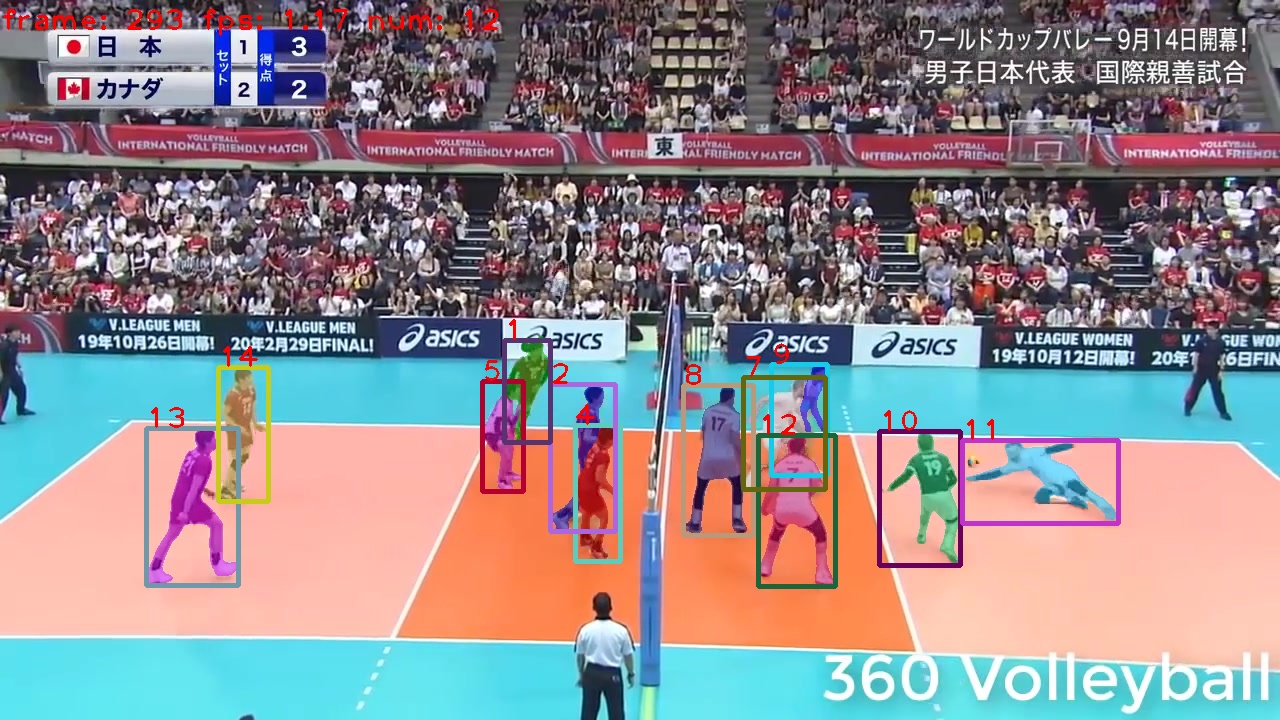}
\\
Baseline \\
\includegraphics[height=3.0cm]{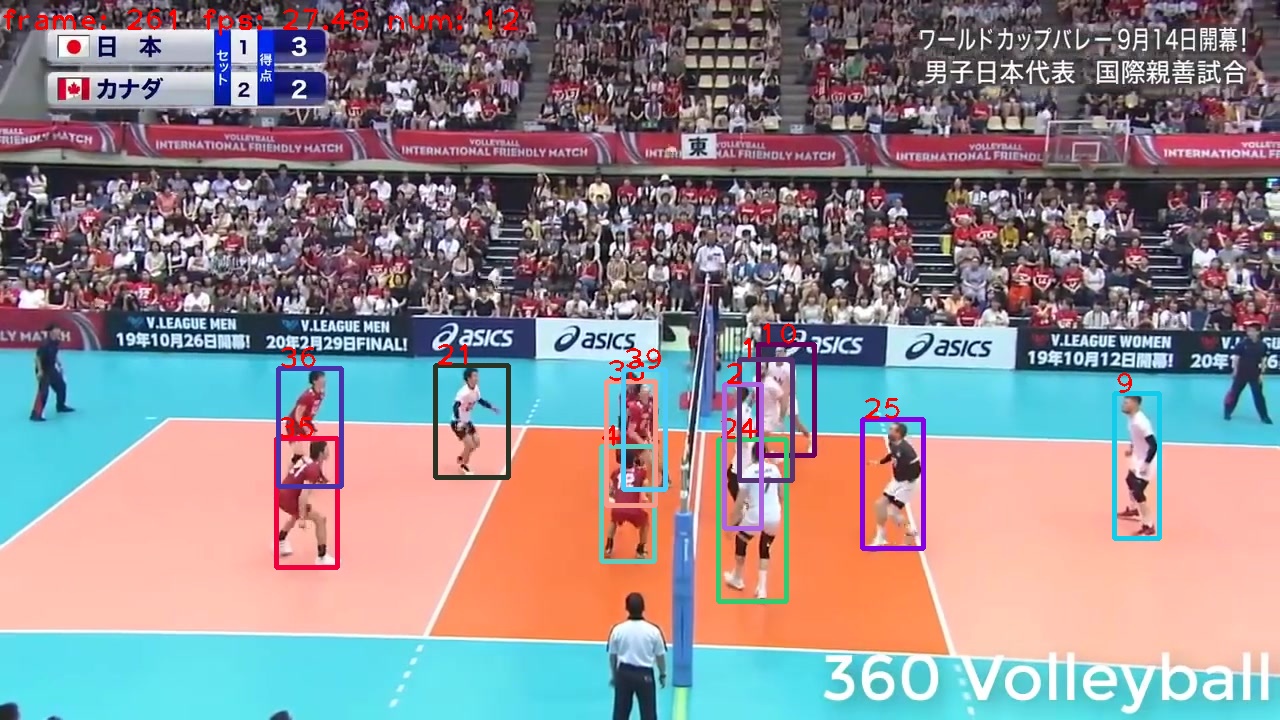}
\includegraphics[height=3.0cm]{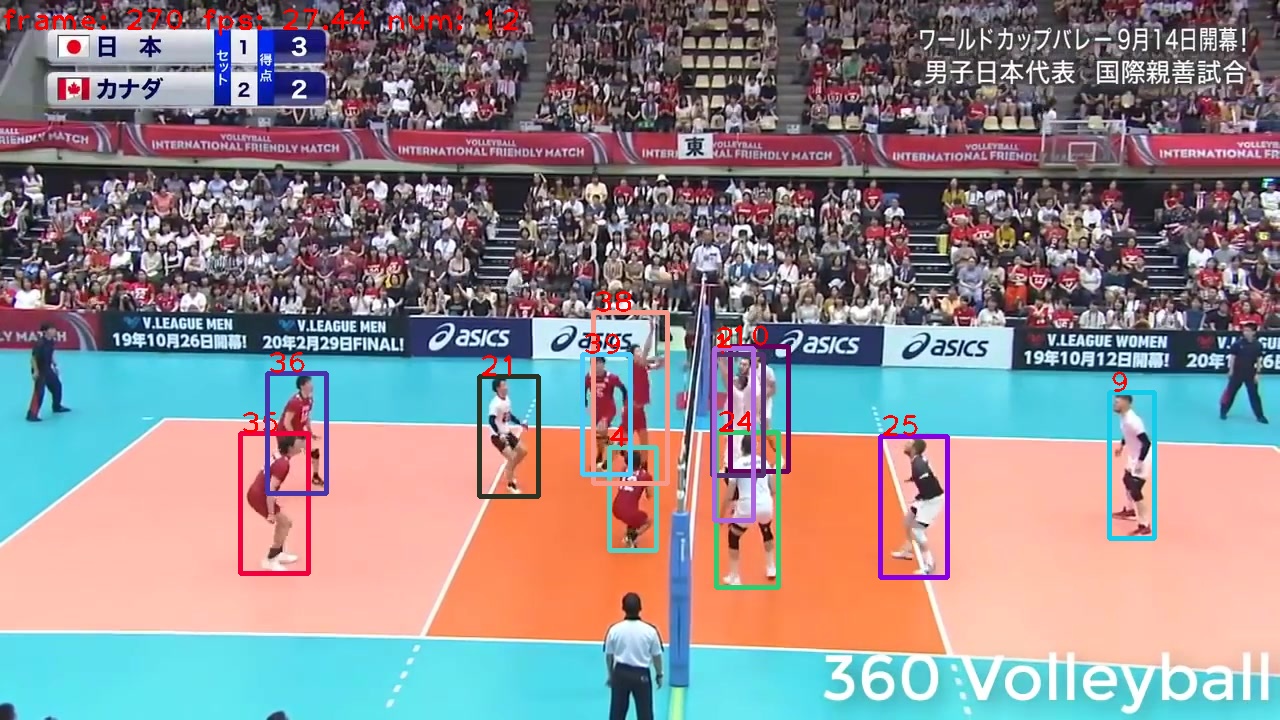}
\includegraphics[height=3.0cm]{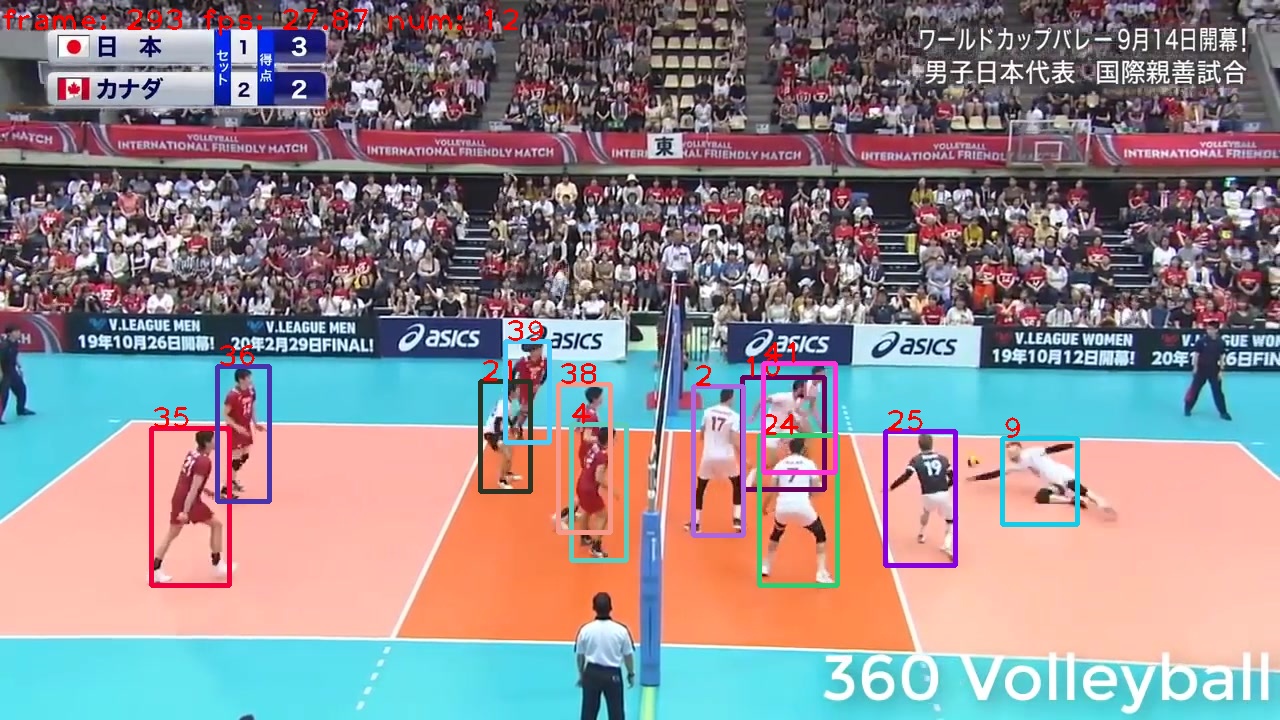}
\end{tabular}
\caption{
Full input and output frames corresponding to Fig. 4 from the main paper. Input image data from~\cite{sportsmot_ref}.
}
% \vspace*{-0.3cm}
\label{fig:full_volley_comp}
\end{figure*}

\begin{figure*}
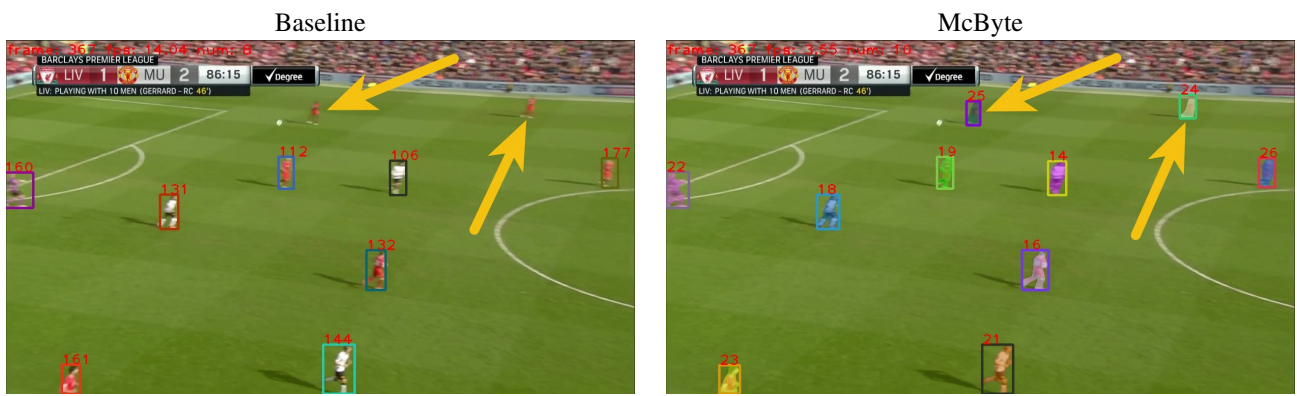

\centering
\begin{tabular}{cc}
Baseline & McByte \\
\includegraphics[height=4.7cm]{images/bytetrack_000367.jpg} &
\includegraphics[height=4.7cm]{images/mcbyte_extra_1_000367.jpg}
\end{tabular}
\caption{
Larger version of Fig. 5 from the main paper. Input image data from~\cite{sportsmot_ref}.
}
% \vspace*{-0.3cm}
\label{fig:full_football_comp}
\end{figure*}

We provide full frame inputs and outputs of the examples used in the main paper, see \cref{fig:full_basket_comp,fig:full_volley_comp} in this supplementary material. We also provide a larger version of one figure from the main paper, see ~\cref{fig:full_football_comp}. 

\begin{figure*}
\centering
\begin{tabular}{cccc}
\includegraphics[height=4.5cm]{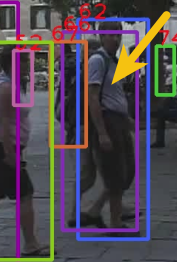}&
\includegraphics[height=4.5cm]{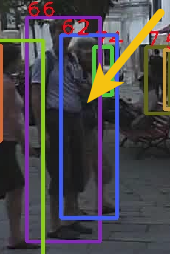}&
\includegraphics[height=4.5cm]{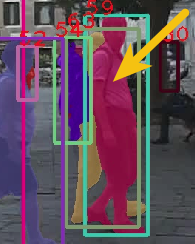}&
\includegraphics[height=4.5cm]{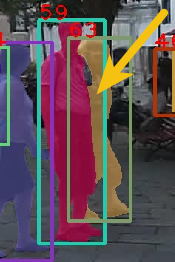}
\\
Frame 459 (baseline)&Frame 475 (baseline)&Frame 459 (McByte)&Frame 475 (McByte)
\end{tabular}
\vspace*{0.3cm}
\caption{Visual output comparison between the baseline and McByte. With the temporally propagated mask guidance, McByte can handle the association of an ambiguous set of bounding boxes 
% and reduce the identity switches
- see the subjects with IDs 59 and 63 on the output of McByte. Input image data from~\cite{mot17_ref}.}
\label{fig:visual_differences_2}
\end{figure*}

\begin{figure*}
\centering
\begin{tabular}{cccc}

\includegraphics[height=3.1cm]{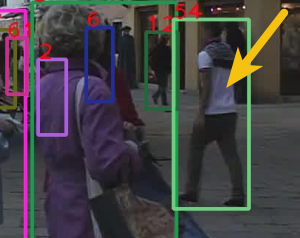}&
\includegraphics[height=3.1cm]{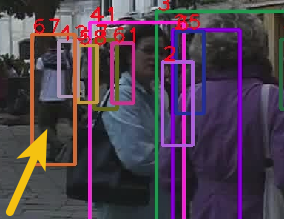}&
\includegraphics[height=3.1cm]{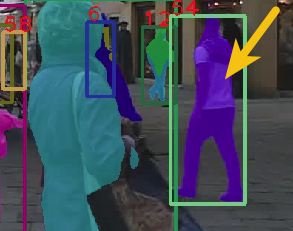}&
\includegraphics[height=3.1cm]{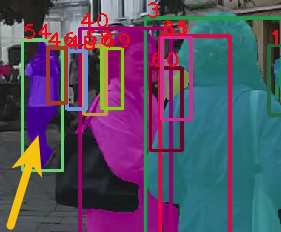}
\\
Frame 319 (baseline)&Frame 401 (baseline)&Frame 319 (McByte)&Frame 401 (McByte)
\end{tabular}
\caption{Visual output comparison between the baseline and McByte. With the temporally propagated mask guidance, McByte can handle longer occlusion in the crowd - see the subject with ID 54 on the output of McByte. Input image data from
~\cite{mot17_ref}.}
% \vspace*{-0.3cm}
\label{fig:visual_differences_1}
\end{figure*}

In the main paper, we discuss that McByte can handle challenging scenarios due to the temporally propagated mask signal used in the controlled manner as an association cue (\cref{sec:method_mask_use}). \cref{fig:visual_differences_2} in this supplementary material shows another example of our method handling association of ambiguous boxes, improving over the baseline. \cref{fig:visual_differences_1} shows an example of our method handling longer occlusions in the crowd.

\section{The running speed and heaviness of mask}
\label{sec:app_speed_and_heaviness}

The running speed of McByte oscillates around 3-5 FPS over the datasets examined~\cite{sportsmot_ref, dancetrack_ref,soccernet-tracking2022_ref,mot17_ref} on a single A100 GPU. It is more costly compared to the baseline~\cite{bt_ref} and other derived methods, but McByte is more reliable - it generalizes well on 4 different datasets and we do not tune it per dataset or per sequence. We believe that it is a good trade-off. Mask-based tracking is a promising concept and we believe it will be further optimized in the community.


 \clearpage

\end{appendices}

\end{document}